\documentclass{article}

\usepackage{hyperref}

\usepackage[accepted]{icml2024}

\usepackage{times}
\usepackage{latexsym}

\usepackage[T1]{fontenc}

\usepackage[utf8]{inputenc}
\usepackage{microtype}
\usepackage{inconsolata}
\usepackage{amsbsy}
\usepackage{pifont}

\usepackage{blkarray}

\usepackage[procnames]{listings}

\definecolor{keyCode}{RGB}{255,0,90}
\definecolor{comments}{RGB}{0,0,113}
\definecolor{red}{RGB}{160,0,0}
\definecolor{green}{RGB}{0,100,0}

\usepackage[export]{adjustbox}

\newcommand\myeq{\mkern1.5mu{=}\mkern1.5mu}
\usepackage{amsfonts}
\DeclareMathSymbol{\shortminus}{\mathbin}{AMSa}{"39}

\usepackage{xcolor}
\usepackage{siunitx}

\definecolor{Gray}{RGB}{128,128,128}
\usepackage{color, colortbl}
\usepackage{booktabs}
\usepackage{array}
\usepackage{makecell}

\usepackage{caption}
\usepackage{subcaption}
\usepackage[thinlines]{easytable}

\newlength{\starlen}
\usepackage{amsmath}
\usepackage{amsthm}
\usepackage{amsfonts}
\usepackage{amssymb}
\usepackage{mathtools}

\newcommand{\sigmoid}{\mathrm{sigmoid}}

\newcommand{\vtheta}{{\boldsymbol \theta}}

\newcommand{\xx}{\mathbf{x}}

\newcommand{\qq}{\mathbf{q}}
\newcommand{\kk}{\mathbf{k}}
\newcommand{\vv}{\mathbf{v}}
\newcommand{\oo}{\mathbf{o}}

\mathchardef\mhyphen="2D

\DeclarePairedDelimiterX{\infdivx}[2]{(}{)}{%
  #1\;\delimsize\|\;#2%
}

\usepackage{todonotes}
\makeatletter
\newcommand*\iftodonotes{\if@todonotes@disabled\expandafter\@secondoftwo\else\expandafter\@firstoftwo\fi}
\makeatother

\definecolor{edolime}{rgb}{0.9,1,0.3}

\definecolor{lightblue}{rgb}{0.7,0.85,1}

\definecolor{orange}{rgb}{1,0.6,0.2}

\usepackage{tikz}
\usetikzlibrary{bayesnet}
\usepackage{graphicx}
\usepackage{booktabs}
\usepackage{tabularx}
\usepackage{tabulary}
\usepackage{colortbl}
\usepackage{xspace}
\usepackage{xcolor}
\newcolumntype{Y}{>{\centering\arraybackslash}X}
\usepackage{multirow}

\newcommand{\RETURN}[1]{{\bf return} #1}

\renewcommand\algorithmiccomment[1]{%
  \hfill\ $\triangleright$\ {#1}%
}

\usepackage[capitalize,noabbrev]{cleveref}

\usepackage{soul}
\DeclareRobustCommand{\hlcyan}[1]{{\sethlcolor{cyan!30}\hl{#1}}}
\DeclareRobustCommand{\hlred}[1]{{\sethlcolor{red!30}\hl{#1}}}

\newcommand{\dmc}{DMC\xspace}

\icmltitlerunning{Dynamic Memory Compression: Retrofitting LLMs for Accelerated Inference}

\begin{document}

\twocolumn[
\icmltitle{Dynamic Memory Compression: Retrofitting LLMs for Accelerated Inference
}
\icmlsetsymbol{equal}{*}
\icmlsetsymbol{edin}{$V$}
\icmlsetsymbol{nvidia}{$Q$}
\icmlsetsymbol{wroc}{$K$}

\begin{icmlauthorlist}
\icmlauthor{Piotr Nawrot}{equal,nvidia,edin}
\icmlauthor{Adrian Łańcucki}{equal,nvidia,wroc}
\icmlauthor{Marcin Chochowski}{nvidia}
\icmlauthor{David Tarjan}{nvidia}
\icmlauthor{Edoardo M. Ponti}{edin}
\end{icmlauthorlist}

\begin{center}
    $^Q$NVIDIA~~~~~$^K$University of Wrocław~~~~~$^V$University of Edinburgh
\end{center}

\icmlcorrespondingauthor{Piotr Nawrot}{\texttt{piotr.nawrot@ed.ac.uk}}

\icmlkeywords{Machine Learning, ICML}

\vskip 0.3in
]

\printAffiliationsAndNotice{\icmlEqualContribution}

\begin{abstract}
Transformers have emerged as the backbone of large language models (LLMs). However, generation remains inefficient due to the need to store in memory a cache of key--value representations for past tokens, whose size scales linearly with the input sequence length and batch size. 
As a solution, we propose Dynamic Memory Compression (\dmc), a method for online key--value cache compression at inference time.
Most importantly, the model learns to apply different compression ratios in different heads and layers.
We retrofit pre-trained LLMs such as Llama 2 (7B, 13B, and 70B) into \dmc Transformers,
achieving up to 7$\times$ throughput increase during auto-regressive inference on an NVIDIA H100 GPU.
\dmc is applied via continued pre-training on a negligible percentage of the original data without adding any extra parameters.
\dmc preserves the original downstream performance with up to 4$\times$ cache compression, outperforming up-trained grouped-query attention (GQA) and key--value eviction policies (H$_2$O, TOVA). GQA and \dmc can be even combined to obtain compounded gains. Hence, \dmc can serve as a drop-in replacement for KV caching in existing LLMs to fit longer contexts and larger batches within any given memory budget.

\end{abstract}

\section{Introduction}

Transformer Large Language Models (LLMs) are the state of the art in generative and conversational AI \citep{touvron2023llama,jiang2023mistral}. Their deployment, however, is curtailed in part by their inefficiency. This is not only due to the quadratic complexity of attention layers \citep{Bahdanau2014NeuralMT,Vaswani2017AttentionIA}: during generation,
Transformers store the keys and values of past tokens in memory to avoid recomputing them multiple times. 
Since this key--value (KV) cache grows linearly with the sequence length and batch size, generation with Transformers quickly becomes prohibitive due to the excessive memory load. This issue emerges even more clearly with long-context generation (e.g., in dialogues and stories) or when serving large numbers of user queries.

\begin{figure}[t]
    \centering
    \begin{subfigure}[b]{1.0\columnwidth}
      \centering
      \includegraphics[width=0.7\columnwidth]{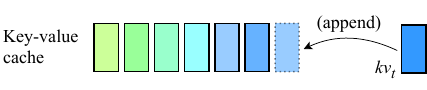}
      \caption{Regular key--value cache with items $kv_i$ depicted as boxes. New items are always appended.}
    \end{subfigure}
    \begin{subfigure}[b]{1.0\columnwidth}
      \centering
      \includegraphics[width=0.7\columnwidth]{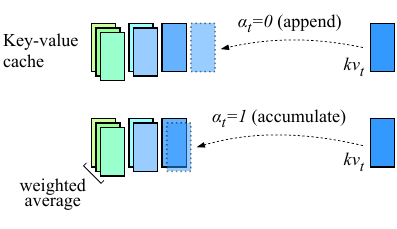}
      \caption{Dynamic Memory Compression (DMC) chooses whether to accumulate or append current items, resulting in a smaller key--value cache.}
    \end{subfigure}
    \caption{Key--value cache update mechanisms.}
    \label{fig:dmc-tldr}
\end{figure}

A widespread solution to increase the memory efficiency of Transformers during inference is Grouped Query Attention \citep[GQA;][]{Ainslie2023GQATG,Shazeer2019FastTD}, where the number of key and value heads is reduced by sharing each of them among multiple query heads.
Alternatively, the length of the KV cache can be shortened through token merging \citep{Zhang2018AcceleratingNT,Liu2018GeneratingWB,Bolya2022TokenMY} or cache eviction policies \citep{Zhang2023H2OHO,Oren2024TransformersAM}. Nevertheless, these methods often pay the price of a severe degradation in downstream performance. On the other hand, hardware/IO-aware \citep{dao2022flash,Kwon2023EfficientMM} and sub-quadratic algorithms for attention \citep{Beltagy2020Longformer, Choromanski2020RethinkingAW} do not alleviate the memory load of the KV cache. 

In our work, we aim to adaptively compress the KV cache of LLMs, retaining their performance while reducing their memory load. To this end, we propose Dynamic Memory Compression (\dmc). As shown in \cref{fig:dmc-tldr}, during every time step, \dmc decides whether to append the current key and value representations to the cache or to perform a weighted average of them with the top-most item in the cache. Note that, for an appropriate prior over compression decisions, the memory can grow sub-linearly in \dmc, which therefore falls halfway between vanilla Transformers and state space language models \citep{fu2023hungry,gu2023mamba}, where memory size is constant.

In our experiments, we equip pre-existing LLMs---such as Llama 2 \citep{touvron2023llama} 7B, 13B, and 70B---with \dmc by retrofitting them on a negligible percentage of the original pre-training data (\textasciitilde2\% for 2$\times$ compression, and \textasciitilde8\% for 8$\times$ compression) and without adding any extra parameters to the original LLM. We evaluate our \dmc models on a series of downstream tasks such as MMLU \citep{hendrycks2021measuring} for factuality, QA datasets for common-sense reasoning, and HumanEval \citep{Chen2021EvaluatingLL} for code. We find that \dmc LLMs retain a downstream performance similar to the original LLM, whereas baselines---such as GQA, H$_2$O, and TOVA---incur significant degradation at high compression ratios. Finally, we show that DMC can be hybridized with GQA such that their compression ratios are compounded. For Llama 2 70B, which is pre-trained with GQA $8\times$, \dmc $2\times$ achieves a total compression of $16\times$.

We verify that KV cache compression translates into more efficient generation in practice. We measure that \dmc $4\times$ increases the inference throughput between 350\% and 390\% for Llama 2 7B and 13B on NVIDIA H100 or A100 GPUs without loss in performance. For higher compression, such as \dmc $8\times$, we observe inference throughput gains up to 700\% with \textasciitilde5\% relative MMLU performance drop for Llama 2 7B and 13B. In fact, \dmc allows us to fit larger batches and longer sequences into a given memory budget. Finally, compression schemata learned by \dmc provide insights into the internal structure of the LLMs, revealing a preference for compressing heads in higher layers.

\section{Background}

\subsection{Multi-Head Self-Attention}
\label{ssec:mhsa}
Let $X=(\xx_1, \ldots, \xx_n) \in \mathbb{R}^{n \times d}$ denote the input sequence of hidden states of a Transformer layer, where $n$ stands for the number of tokens in a sequence, and $d$ for the hidden state dimension. A Multi-Head Self-Attention (MHSA) layer 
divides the embeddings into $n_h$ different heads. Afterward, the self-attention process is applied to each head separately. 
This enables the model to focus on different parts of the input, capturing various types of relationships in the data.
For each head $h$, different weight matrices $W_q^h, W_k^h, W_v^h \in \mathbb{R}^{d/n_h \times d/n_h}$ are used to project the input sequence into queries $Q^h$, keys $K^h$, and values $V^h$:
\begin{align}
Q^h &= (W_q^h\xx_1, \ldots, W_q^h\xx_n) \\
K^h &= (W_k^h\xx_1, \ldots, W_k^h\xx_n) \\
V^h &= (W_v^h\xx_1, \ldots, W_v^h\xx_n).
\end{align}
The attention scores and outputs for the $i$-th token are then computed as

\begin{equation}
a_{ij}^h = \frac{\exp(\qq_i^{h^\top} \kk_j^h / \sqrt{d_h})}{\sum_{t=1}^i \exp(\qq_i^{h^\top} \kk_t^h / \sqrt{d_h})}, \quad \oo_i^h = \sum_{j=1}^i a_{ij}^h\vv_j^h.
\label{eq:attn-softmax}
\end{equation}
Finally, the outputs from all heads are concatenated and linearly transformed to produce the final output $O \in \mathbb{R}^{n \times d}$:
\begin{equation}
O = (W_o[\oo_1^1, \ldots, \oo_1^{n_h}], \ldots, W_o[\oo_n^1, \ldots, \oo_n^{n_h}])
\end{equation}
where $W_o \in \mathbb{R}^{d \times d}$.

\subsection{KV Caching During Inference}  %
\label{ssec:kv_caching}

In a Transformer decoder, the generation of sequences is auto-regressive: each new token is predicted based on the previously generated ones. To avoid redundant computation, it is common practice to store the keys and values of previously computed tokens in the KV cache. For each time step $i$, only the keys and values for the current token $\xx_i$ are computed whereas those for $\xx_{<i}$ are retrieved from the cache. Thus for each head $h$:
\begin{align}
    K^h &= [K^h_{1:i-1}, W_k^h \xx_i] \\
    V^h &= [V^h_{1:i-1}, W_v^h \xx_i]
\end{align}
Note that this process is not necessary for queries as only the query for the current token $\xx_i$ is needed at each inference time step.

As a consequence, the KV cache grows linearly with each new token. This progressive expansion leads to substantial memory consumption and increased latency, especially for long input sequences. This issue is further exacerbated when serving multiple requests concurrently, as each inference process requires its own KV cache, significantly straining the system's resources.

\subsection{Memory-Bound and Compute-Bound Operations}
\label{ssec:memory_compute_bound}
Every operation performed with a GPU accelerator, such as General Matrix Multiply (GEMM), is either memory-bound or compute-bound. In the former case, the overall runtime is dominated by high bandwidth memory (HBM) access, while in the latter by the actual computations. Auto-regressive generation with Transformer LLMs, where the sequence length for every forward pass is $n=1$, tends to be memory-bound rather than compute-bound. The vast majority of a forward pass is spent either processing linear layers (in MHSA, Feed-Forward, and output vocabulary projection) or calculating attention scores and outputs from \cref{eq:attn-softmax}. For linear layers, the ratio of FLOPS to memory accesses improves as the batch size increases, and more FLOPS are performed with the set of layer weights retrieved from the HBM. Eventually, with a large enough batch size, linear layers become compute-bound. On the other hand, for the calculation of \cref{eq:attn-softmax} inside MHSA layers during auto-regressive inference, the ratio of FLOPS to input size remains constant, and MHSA layers are memory-bound regardless of the batch size. It follows that for those layers, latency scales linearly with the size of the KV cache.

\section{Method: Dynamic Memory Compression}
In this work, we tackle the problem of reducing the size of the KV cache, which brings two immediate benefits \citep{Zhang2023H2OHO}: 1) it lowers the latency of auto-regressive generation, and 2) improves GPU utilization by allowing for larger batch sizes and sequence lengths, leading to increased throughput. In this section, we introduce Dynamic Memory Compression (\dmc), a simple and inexpensive method for online compression of the KV cache at inference time. In what follows, we first describe the inference-time operation of \dmc, which constitutes our end goal. Next, we illustrate how to teach a pre-trained LLM such behavior through short, continued pre-training.

\subsection{Inference}
\label{ssec:inference}

Consider the forward pass of an attention layer during auto-regressive inference (\cref{ssec:mhsa}).
In a vanilla Transformer, at every time step $t$, both $\kk_t$ and $\vv_t$ are appended to the KV cache (\cref{ssec:kv_caching}). In \dmc, on the other hand, the KV cache update procedure is different, as detailed in \cref{alg:dmc-update}.
First, a decision variable $\alpha_t\in\{0,1\}$ and importance variable $\omega_t\in[0, 1]$ are predicted.
In order to avoid adding new parameters, we reuse the first neuron from $\kk_t$ and $\qq_t$, respectively, to extract the two scores:\footnote{Note that this choice is arbitrary as we could use any two neurons from $\{ \qq_t, \kk_t, \vv_t \}$.}
\begin{equation}
    \label{eq:reuse}
    \alpha_t \myeq \lfloor \text{sigmoid}(\kk_t[0]) \rceil,\  \omega_t \myeq \text{sigmoid}(\qq_t[0]).
\end{equation}
where sigmoid normalizes both scores into a range $[0, 1]$. At inference time, the decision score $\alpha_t$ is obtained by rounding to the closest integer to make this variable binary.

\begin{algorithm}[ht]
\caption{Single-head KV cache update with Dynamic Memory Compression (DMC)}\label{alg:dmc-update}
\begin{algorithmic}[1]
\REQUIRE $K, V, \qq_t, \kk_t, \vv_t$
\STATE $\alpha_t \gets \text{round}(\text{sigmoid}(\kk_t[0]) )$ 
\STATE $\omega_t \gets \text{sigmoid}(\qq_t[0])$
\STATE $\kk_t[0], \qq_t[0] \gets 0, 0$
\IF[\hlcyan{\textsc{accumulate}}]{$\alpha_t = 1$}
\STATE $z_t \gets z_{t-1} + \omega_t$
\STATE $K \gets [K_{1:l-1}, (K_l z_{t-1} + \kk_t \omega_t)/z_t]$
\STATE $V \gets [V_{1:l-1}, (V_l z_{t-1} + \vv_t \omega_t)/z_t]$
\ELSE[\hlred{\textsc{append}}]
\STATE $z_t \gets \omega_t$
\STATE $K \gets [K_{1:l}, \kk_t]$
\STATE $V \gets [V_{1:l}, \vv_t]$
\ENDIF
\STATE\RETURN $K, V, \qq_t, \kk_t$
\end{algorithmic}
\end{algorithm}

Based on $\alpha_t$, a decision is made whether KV representations $\kk_t$ and $\vv_t$ are \hlred{appended} to the cache or \hlcyan{accumulated} with its last element (\cref{fig:dmc-tldr}).
In particular, for accumulation, we perform a weighted average based on the predicted importance score $\omega_t$ for the current token and the running sum of importance scores $z_{t}$ for all the tokens since the last time step $\alpha = 0$ was predicted.

In effect, the $\alpha$ variable segments the input sequence: each decision determines if the current segment should continue ($\alpha=1$) or a new segment should be opened ($\alpha = 0$).
As a result, after the update, the cache length for DMC is $l =\sum_{t=1}^n (1-\alpha_t)\leq n$, whereas in vanilla Transformers it is always $l=n$.
In what follows, we will refer to the ratio $n/l$ between the length $n$ of an uncompressed cache and the compressed length $l$ as the Compression Ratio (CR).

Finally, multi-head self-attention is calculated similarly to vanilla Transformers but using the compressed KV cache sequences. %
\cref{alg:dmc-update} is applied to every MHSA layer and head independently. Hence, the corresponding KV sequences might have different lengths.
Note that \cref{alg:dmc-update} can be implemented efficiently without the if-then-else statement conditioned on $\alpha_t$, instead multiplying the previous ${\kk}_i$, ${\vv}_i$ and $z_i$ by $\alpha_t$ like in \cref{eq:partial-accumulation}.

\subsection{Training}
\label{method:uptraining}

The \dmc inference algorithm switches between accumulating and appending tokens to the KV cache. In order to endow LLMs with \dmc, we continue pre-training them on a negligible amount of their pre-training data, gradually increasing the compression ratio towards a target. However, this poses serious challenges. First, we opt for end-to-end learning via gradient descent, which requires a continuous relaxation of the decision variables. As a result, we have to define an operation for KV cache updating when $0 < \alpha < 1$, resulting in partly aggregated, partly accumulated key and value states. Second, to avoid training--inference mismatch, we must simulate the \dmc behavior at inference time while parallelizing training across a sequence of tokens. As a consequence the length of $K$ and $V$ is {\it not} reduced through compression during training; rather, all intermediate states of keys and values are explicitly kept in the sequence and an auxiliary, gradually discretized mask regulates interactions between queries and keys. We elaborate on our solutions to these challenges below.

\paragraph{Gradient Estimation for Discrete Decisions}
The decision whether to accumulate or append at inference time is a discrete one; however, rounding $\alpha_t = \sigmoid(\kk_t[0])$ to the nearest integer for training would result in an operation with zero or undefined gradients. Hence, we resort to stochastic reparametrization of the decision variable during training
\begin{equation} \label{eq:gumbel_sigmoid}
    \alpha_t \sim \text{Gumbel-sigmoid}(\kk_t[0] - c, \tau) \in[0,1],
\end{equation}
where $\tau$ is the temperature\footnote{Low temperatures sharpen $\alpha_t$ into almost-discrete values which accurately mimics inference behavior.} and $c$ is a constant subtracted so that every $\alpha \approx 0$ at training step $0$. %
This ensures that \dmc initially performs no compression and starts the training behaving just like the vanilla Transformer.

\paragraph{Partial accumulations}
As we relax our discrete decisions, we must now define a mechanism to update the KV cache that generalizes \cref{alg:dmc-update} to continuous $\alpha$. Hence, we define partially accumulated states for $\alpha\in[0,1]$ as:
\begin{equation} \label{eq:partial-accumulation}
\begin{aligned}
z_0 & \gets \omega_0,\quad  & z_i & \gets z_{i-1} \alpha_i + \omega_i, \\
{\kk}_0 & \gets \kk_0,\quad & \kk_i & \gets \frac{\alpha_i\kk_{i-1}z_{i-1} + \kk_i\omega_i}{z_i}, \\
{\vv}_0 & \gets \vv_0,\quad & \vv_i & \gets \frac{\alpha_i\vv_{i-1}z_{i-1} + \vv_i\omega_i}{z_i}.
\end{aligned}
\end{equation}
Note that when $\alpha\in\{0,1\}$, \cref{eq:partial-accumulation} defaults to \cref{alg:dmc-update}.

\paragraph{Intermediate compression steps}
Aside from key and value computations shown in \cref{eq:partial-accumulation}, the rest of the forward pass can be performed in parallel for all tokens in the sequence. Nonetheless, this creates a mismatch between training and evaluation, since during training all intermediate states of keys and values are accessible in self-attention.

\begin{figure}[t]
    \centering
    \begin{subfigure}[b]{0.65\columnwidth}
      \centering
      \includegraphics[width=1.05\columnwidth]{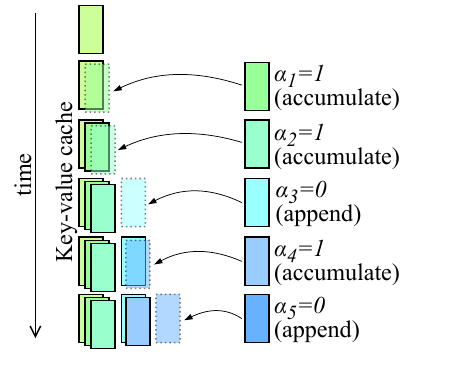}
    \end{subfigure}
    \begin{subfigure}[b]{0.33\columnwidth}
      \centering
      \includegraphics[width=1.0\columnwidth]{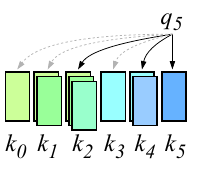}
    \end{subfigure}
    \caption{An example of KV cache growth in DMC during inference (left). During training (right), we retain all intermediate states seen during inference and gradually block access to some of them (gray arrows)}
    \label{fig:dmc_cache_unrolled}
\end{figure}

To illustrate this issue, consider the example of a KV cache during DMC inference shown in \cref{fig:dmc_cache_unrolled} for the sequence of decision scores $\alpha_{1:5}=(1, 1, 0, 1, 0)$ (importance scores $\omega$ have been omitted for clarity). Because the last element of the KV cache changes in-place at every time step, the future elements should not have access to its old contents (\cref{fig:dmc_cache_unrolled}, right).
In order to properly simulate this inference-time evolution of the KV cache during training, we keep all unrolled intermediate KV cache items. However, in lieu of an auto-regressive `causal' mask, we use an additive mask based on the sequence of $\alpha$ values to modify the attention scores 
$a^h_{ij}$ from \cref{eq:attn-softmax}, which is shown in \cref{tab:example_masking}. %
As $\alpha$ values are increasingly pushed to almost-discrete states by the Gumbel estimator and the low-temperature setting, this strengthens the interactions of queries with the last state of every key--value segment (i.e., the only one accessible during inference), and weakens those with the previous states, which are discarded during inference.\footnote{
See \cref{app:masking_details} for details on the mask implementation.}
In fact, when $\alpha\in\{0,1\}$, the matrix is filled with either $0$ or $-\infty$ values, and exactly corresponds to the inference-time query-to-key attendance pattern.

\begin{figure}
  \resizebox{1.0\columnwidth}{!}{%
  $\arraycolsep=1.0pt\def\arraystretch{1.2}
  \begin{blockarray}{cccccc}
   & {\bf k_0} & {\bf k_1} & {\bf k_2} & \dots & {\bf k_n} \\
  \begin{block}{c[ccccc]}
    {\bf q_0} & 0 & -\infty & -\infty & \dots  & -\infty \\
    {\bf q_1} & \log (1 \shortminus \alpha_1) & 0 & -\infty  &   & -\infty \\
    {\bf q_2} & \log (1 \shortminus \alpha_1)  & \log (1 \shortminus \alpha_2)  &  0 &  & -\infty \\
    \vdots & \vdots & & & \ddots & \vdots \\
    {\bf q_n} & \log (1 \shortminus \alpha_1)  & \log (1 \shortminus \alpha_2)  &  \log (1 \shortminus \alpha_3) &  \dots & 0 \\
  \end{block}
  \end{blockarray}
  $}
\caption{Additive mask applied during training to the normalized attention scores in order to block queries from attending intermediate KV states (as well as future states).}
\label{tab:example_masking}
\end{figure}

\paragraph{Training objective}
The model is incentivized to compress the KV cache to a certain CR, and thus increase the predicted $\alpha$ values. Instead of matching the desired rate for each append-or-accumulate decision $\alpha$, we calculate a \textit{global} one-sided loss as the difference between the sum of all decisions and the expected sum of KV tokens across all layers $l$, heads $h$ and time steps $t$ under the desired Compression Ratio ($\text{CR}$), normalized by $(n_l n_h n)$:
\begin{align} \label{eq:cr_loss}
    \ell_{\text{CR}} = \frac{1}{n_l n_h n} \text{max}\left(0, \sum_{l=1}^{n_l} \sum_{h=1}^{n_h} \sum_{t=1}^{n} (1 - \alpha_{lht}) - \frac{n_l n_h n}{\text{CR}}\right).
\end{align}

The $\ell_{\text{CR}}$ loss is added to the language modeling loss term 
$ \ell_{\text{LM}} = - \sum_{t=1}^n \log p_\vtheta(x_t \mid \xx_{<t}) $, with the final objective of the training being:
\begin{equation}
    \text{arg min}_\vtheta \, \ell_{\text{LM}} + \ell_{\text{CR}}.
\end{equation} 

Importantly, the training procedure is designed for slowly ramping up the target $\text{CR}$ and saving ready-to-use DMC checkpoints along the way. This is possible because all hyperparameters, like Gumbel-sigmoid sampling temperature and learning rate, are not decayed and remain constant throughout training. A practical use case of this DMC property is to produce a series of DMC checkpoints with different CRs within a single run, and then choose the one with the desired efficiency--performance trade-off. 

\subsection{Practical Considerations}
\paragraph{Storing a variable-length KV cache in memory} 
\dmc allows every head to learn a custom compression, which results in KV cache sequences with variable lengths across heads. This poses difficulties in storing these sequences efficiently in an $n$-dimensional tensor, considering that they will be extended during auto-regressive generation by uneven amounts of tokens. However, such sequences can be easily stored in memory with little overhead using PagedAttention \citep{Kwon2023EfficientMM}, where new pages are allocated on-demand for every head separately. In \cref{sec:latency} we present latency and throughput measured with an implementation based on FlashAttention \citep{dao2022flash} and PagedAttention.

\paragraph{Window grouping approximation}
\label{method:window}
The calculation of partial accumulations, during training of DMC models \cref{eq:partial-accumulation}, for a sequence of $n$ tokens requires $O(n)$ sequential steps, therefore considerably slowing down the training. In order to improve the time complexity, we calculate \cref{eq:partial-accumulation} at every time step $t$ independently over a short window of the last $w$ elements up to time $t$ (e.g., $w = 12$). This enables us to reduce the span of computation to $O(w)$, provided that at least $n$ threads can execute the computation in parallel for each of the $n$ positions. During inference, this speed-up also applies to the prompt phase. However, the sliding window comes at a disadvantage as it needs to be cached during inference for heads that tend to accumulate more than $w$ tokens.

\section{Experimental Setup}

\paragraph{Baselines}

In our experiments, we evaluate strategies to retrofit a state-of-the-art Large Language Model (LLM), Llama 2 \citep{touvron2023llama} \footnote{Obtained from \href{https://huggingface.co/meta-llama}{https://huggingface.co/meta-llama}.}, into a more efficient model across various sizes: 7B, 13B, and 70B. In addition to comparing the downstream performance of \dmc with the original model, we also use Grouped Query Attention (GQA) as a main baseline, as it constitutes the most widespread strategy to ensure KV cache efficiency \cite{jiang2023mistral}.

We also compare DMC with two KV cache eviction policies that do not require model retrofitting: Token Omission Via Attention \citep[TOVA;][]{Oren2024TransformersAM} and Heavy-Hitter Oracle \citep[H$_2$O;][]{Zhang2023H2OHO}. H$_2$O keeps in memory a fixed window of the most recent tokens, as well as additional heavy-hitter (H$_2$) tokens. H$_2$ tokens are chosen dynamically: specifically, they correspond to the tokens with the highest aggregated attention scores throughout the sequence. On the other hand, TOVA retains the top-k tokens based on the attention weights of the last token only. This means that for a given attention layer, at each decoding step, the token with the lowest attention score is omitted.

\paragraph{Checkpoint adaptation} 
To equip the original Llama 2 LLM with GQA, we perform the standard checkpoint conversion proposed by \citet{Ainslie2023GQATG}: the key and value projection matrices are split by head. Then the resulting sub-matrices corresponding to heads in the same group are merged via averaging. Note that this results in a fixed compression during training.

As for \dmc, we avoid the introduction of new parameters by re-purposing the first dimension from both the $\qq_t$ and $\kk_t$ representations, in order to predict segmentation decisions $\alpha_t$ and importance scores $\omega_t$. 
Setting $\qq_t$ and $\kk_t$ to zero triggers a significant increase in language modeling loss, by disrupting the attention scores. To counteract this spike in loss, we pre-train the model to disregard the first dimension of $\qq_t$ and $\kk_t$ in the attention calculations. Specifically, we load the pre-trained weights and up-train the model for 1 billion tokens (250 steps), annealing the values of $\qq_t$ and $\kk_t$ to 0 according to the following formula:

\begin{equation}
\begin{aligned}
    \qq_{t}[0] &\gets \qq_{t}[0] \times (1 - (t / n_t)) \\
    \kk_{t}[0] &\gets \kk_{t}[0] \times (1 - (t / n_t))
\end{aligned}
\end{equation}
where $t$ is the current step and $n_t=250$. After this initial phase, which allows the model to ignore the first dimension of keys and values for attention calculations, we start the main retrofitting phase, where the model learns to compress the KV representations.

\paragraph{Training hyperparameters}
We strictly follow the training hyperparameters outlined by \citet{touvron2023llama}. We employ the AdamW optimizer with parameters \(\beta_1 = 0.9\), \(\beta_2 = 0.95\), and \(\epsilon = 1e-5\), in conjunction with a weight decay of $0.1$ and gradient clipping of $1.0$. The batch size is $1024$ with a sequence length of $4096$.
We apply a constant learning rate identical to the final rate from the original Llama 2 pre-training phase: \num{3e-5} for the 7B and 13B models, and \num{1.5e-5} for the 70B model. We set the constant from \cref{eq:gumbel_sigmoid} as $c = 5$ which in practice results in $\alpha_t = 0.0067$. Empirically, $c = 5$ is a high enough value so that we do not experience a spike in language modeling loss at the start, yet low enough to be easily changed by learning $\qq_t[0]$ and $\kk_t[0]$ through gradient optimization. Finally, we set the window size (\Cref{method:window}) to 12, and keep the Gumbel-sigmoid temperature $\tau$ constant at 0.1 for the entire training. We do not perform any exhaustive searches for these values: we believe that the DMC retrofitting procedure is robust to a wide range of sensible hyperparameter choices.

\paragraph{Training schedule}
The volume of data for continued pre-training of \dmc is contingent on the targeted KV cache compression ratio; a larger CR necessitates more data. We use a linear training schedule with 24B, 72B, 120B, and 168B tokens for training to $2\times$, $4\times$, $6\times$, and $8\times$ compression, respectively. In \Cref{app:retrofitting_ablations} we include an ablation where we use a schedule with twice less data.

We discovered that the right retrofitting strategy is crucial for \dmc: starting the training without compression helps to preserve the original perplexity. Any significant increase in perplexity, even if recovered during continued pre-training, prevents the model from regaining its performance on downstream tasks (see the ablations in \cref{app:retrofitting_ablations}). The target CR is linearly increased from $1\times$ to $8\times$ for the 7B and 13B models, and from $1\times$ to $2\times$ for the 70B model.\footnote{For Llama 2 70B, we do not up-train to $4\times$ because this LLM is already pre-trained with GQA $8\times$.}

Upon achieving the target compression ratio, we initiate a final solidifying phase wherein we: 1) continue up-training for an additional 8B tokens, 2) maintain a fixed compression ratio, and 3) implement a cosine learning rate schedule, annealing it down to 10\% of the initial value. This phase aims at stabilizing the model with a specific, fixed compression ratio. Resource-wise, all retrofitting stages require roughly 8k, 16k, and 28k GPU hours of NVIDIA H100 for Llama 2 7B, 13B, and 70B respectively.

\paragraph{Evaluation}

Following \citet{touvron2023llama}, we evaluate models on a series of downstream tasks, including MMLU \citep{hendrycks2021measuring} for factuality, HumanEval \citep{Chen2021EvaluatingLL} for Python code generation, and several question-answering datasets for common-sense reasoning: PIQA \citep{bras_Gao_Choi_2020}, BoolQ \citep{clark-etal-2019-boolq}, Arc-C and Arc-E \citep{clark2018think}, HellaSwag \citep{zellers-etal-2019-hellaswag}, and WinoGrande \citep{Sakaguchi2019AnAW}. We report the 5-shot performance on MMLU, average pass@1 scores for HumanEval, and average 0-shot performance on common-sense benchmarks (CS-QA). For pass@1 scores we use a temperature of 0.1 and nucleus sampling \citep{Holtzman2019TheCC} with top-p $=$ 0.95.

We adapted TOVA and H$_2$O to our codebase based on publicly released code. For a given CR, the total budget for both policies is calculated as $(1/\text{CR}) \times n$ for MMLU and CS-QA tasks, where $n$ is the input length. For HumanEval, which instead involves generating more than one token, the initial budget is also $(1/\text{CR}) \times n$ but then increases dynamically as we generate the answer. For H$_2$O, the budget is split equally between the local window and heavy-hitter tokens.

\section{Results}

\begin{table}[th!]
\centering
\resizebox{1.0\columnwidth}{!}{%
\begin{tabular}{c c c r r r}
\toprule
\textbf{Scale} & \textbf{Method} & \textbf{CR} & \textbf{MMLU} & \textbf{CS-QA} & \multirowcell{2}{\textbf{Human}\\ \textbf{Eval}} \\ 
& \\
\midrule
\multirow{17}{*}{7B} & -- & -- & 44.6 & 70.5 & 14.0 \\ 
\cmidrule{2-6}
 & GQA    & \multirow{4}{*}{2$\times$} & 39.8 & 68.9 & 12.8 \\
 & H$_2$O & & \textbf{45.2} & 67.5 & 9.8 \\
 & TOVA & & 44.9 & 70.0 & 6.1 \\
 & DMC & & \textbf{45.2} & \textbf{70.8} & \textbf{15.2} \\
 \cmidrule{2-6}
 & GQA & \multirow{4}{*}{4$\times$} & 34.7 & 68.3 & 14.0 \\
 & H$_2$O & & 41.1 & 56.8 & 4.9 \\
 & TOVA & & 43.4 & 64.2 & 1.8 \\
 & DMC & & \textbf{43.9} & \textbf{70.2} & \textbf{16.5} \\
   \cmidrule{2-6}
  & H$_2$O & \multirow{3}{*}{6$\times$} & 36.3 & 51.9 & 0.0 \\
 & TOVA & & 41.1 & 56.1 & 0.0 \\
 & DMC & & \textbf{42.9} & \textbf{70.4} & \textbf{15.9} \\
\cmidrule{2-6}
 & H$_2$O & \multirow{3}{*}{8$\times$} & 32.5 & 48.3 & 0.6 \\
 & TOVA & & 38.7 & 51.0 & 0.0 \\
 & DMC & & \textbf{41.8} & \textbf{70.1} & \textbf{16.5} \\
\midrule
\multirow{17}{*}{13B} & -- & -- & 54.5 & 73.5 & 17.5 \\ 
\cmidrule{2-6}
 & GQA & \multirow{4}{*}{2$\times$}  & 50.2 & 72.7 & 15.9 \\
 & H$_2$O & & 54.1 & 70.3 & 16.5 \\
 & TOVA & & 54.4 & 72.8 & 12.2 \\
 & DMC & & \textbf{54.8} & \textbf{74.2} & \textbf{20.7} \\
 \cmidrule{2-6}
 & GQA & \multirow{4}{*}{4$\times$} & 48.6 & 72.2 & 16.5 \\
 & H$_2$O & & 50.7 & 60.0 & 8.5 \\
 & TOVA & & 53.3 & 67.8 & 2.4 \\
 & DMC & & \textbf{54.2} & \textbf{73.2} & \textbf{22.0} \\
  \cmidrule{2-6}
 & H$_2$O & \multirow{3}{*}{6$\times$} & 45.9 & 54.9 & 2.4 \\
 & TOVA & & 51.7 & 60.0 & 0.0 \\
 & DMC & & \textbf{53.3} & \textbf{73.4} & \textbf{21.3} \\
\cmidrule{2-6}
 & H$_2$O & \multirow{3}{*}{8$\times$} & 41.8 & 50.9 & 1.8 \\
 & TOVA & & 50.3 & 53.4 & 0.0 \\
 & DMC & & \textbf{52.1} & \textbf{73.3} & \textbf{21.3} \\
\midrule
\multirow{4}{*}{70B$^*$} & -- & 8$\times^*$ & 68.8 & 78.0 & 29.6 \\ \cmidrule{2-6}
 & H$_2$O & \multirow{3}{*}{16$\times^*$} & 68.7 & 74.1 & 18.3 \\
 & TOVA & & 68.1 & 77.6 & \textbf{29.9} \\
& DMC & & \textbf{68.8} & \textbf{77.9} & \textbf{29.9} \\
\bottomrule
\end{tabular}
}
\caption{MMLU accuracy, Commonsense Question Answering (CS-QA) accuracy averaged across 6 tasks, and HumanEval Pass@1 for several scales (7B, 13B, and 70B) and compression ratios (CRs; $1\times$, $2\times$, $4\times$, $6\times$, $8\times$) of Llama 2. (*) Unlike the 7B and 13B models, the 70B model was trained with GQA which compresses the KV cache $8\times$. We apply additional $2\times$ DMC compression during retrofitting to obtain a total compression of $16\times$.}
\label{tab:model-performance}
\end{table}

\subsection{Main Results} 
We report the performance of the original LLM (equivalent to $1\times$ CR) and efficient variants (\dmc, GQA, TOVA, and H$_2$O) in \cref{tab:model-performance}. The results for the original LLM are those reproduced in our codebase as described in \cref{app:reproduce_baseline}.

\textbf{\dmc vs Original} First, comparing \dmc with the original LLM, we note that it even \textit{increases} the performance in MMLU and CS-QA at $2\times$ CR for 7B and 13B and in HumanEval for all model scales. We speculate that this is due to the additional up-training steps, which expose LLMs to new examples. For the other combinations of downstream tasks and scales at $4\times$ CR, \dmc incurs negligible degradation: -0.7 in MMLU and -0.3 in CS-QA for 7B, -0.3 in MMLU and CS-QA for 13B. At higher compression ratios of $6\times$ and $8\times$ CR \dmc shows slight degradation but generally stays close to the original performance. For 7B, \dmc exhibits -1.7 in MMLU at $6\times$ CR, and -2.8 at $8\times$ CR. For 13B performance drops are smaller than for 7B, \dmc incurs -1.2 in MMLU and -0.1 in CS-QA at $6\times$ CR, and -2.4 in MMLU and -0.2 in CS-QA at $8\times$ CR. This encouraging finding suggests that \dmc is robust across different model scales even for $8\times$ CR. Overall, the fact that \dmc is in general close or superior to the original performance makes it suitable as a drop-in replacement for vanilla KV caching to achieve higher inference efficiency.

\textbf{\dmc vs GQA} Moreover, \Cref{tab:model-performance} allows for comparing \dmc with GQA for equivalent CRs ($2\times$ and $4\times$). Overall, \dmc surpasses GQA applied \textit{post-hoc} through up-training, in both CRs and in both scales (7B and 13B) where we compare them. For MMLU, the gap widens when we increase the CR. This holds true both
at the smaller 7B scale (from +5.4 for $2\times$ CR to +9.2 for $4\times$ CR) 
and at the larger 13B scale (from +4.6 for $2\times$ CR to +5.6 for $4\times$ CR). For CS-QA and Human-Eval, on the other hand, we observe comparable gains over GQA across CRs and model scales. Notably, \dmc with $8\times$ compression outperforms GQA with $2\times$ compression for both 7B and 13B model scales. These findings illustrate that \dmc should be preferred to GQA for retrofitting LLMs into variants that are more efficient at inference time.

\textbf{\dmc vs Cache Eviction Policies}
Both H$_2$O and TOVA evict tokens from memory according to the post-softmax attention scores, which makes them incompatible with efficient attention implementations like FlashAttention. In fact, this requires materializing the $n^2$-sized tensor of attention scores. As a result, while reducing the KV cache size, these policies may slow down inference. Moreover, comparing the performance of both cache eviction policies with DMC in \cref{tab:model-performance}, it emerges how they are almost comparable at CR 2$\times$ in MMLU and CS-QA; however, they drop significantly in accuracy with a higher CR of 4$\times$, $6\times$, and $8\times$. The gap is even more dramatic for HumanEval pass@1 at all CRs and scales, except only for 70B TOVA. At $6\times$ CR, H$_2$O and TOVA show significant drops in MMLU and CS-QA, with TOVA outperforming H$_2$O. At $8\times$ CR, the performance drops further for both policies, especially in HumanEval, where \dmc consistently shows superior performance.

\begin{figure}[t]
    \vspace{-0.90em}
    \centering
    \hspace{-0.90em}
    \includegraphics[width=1.03\columnwidth]{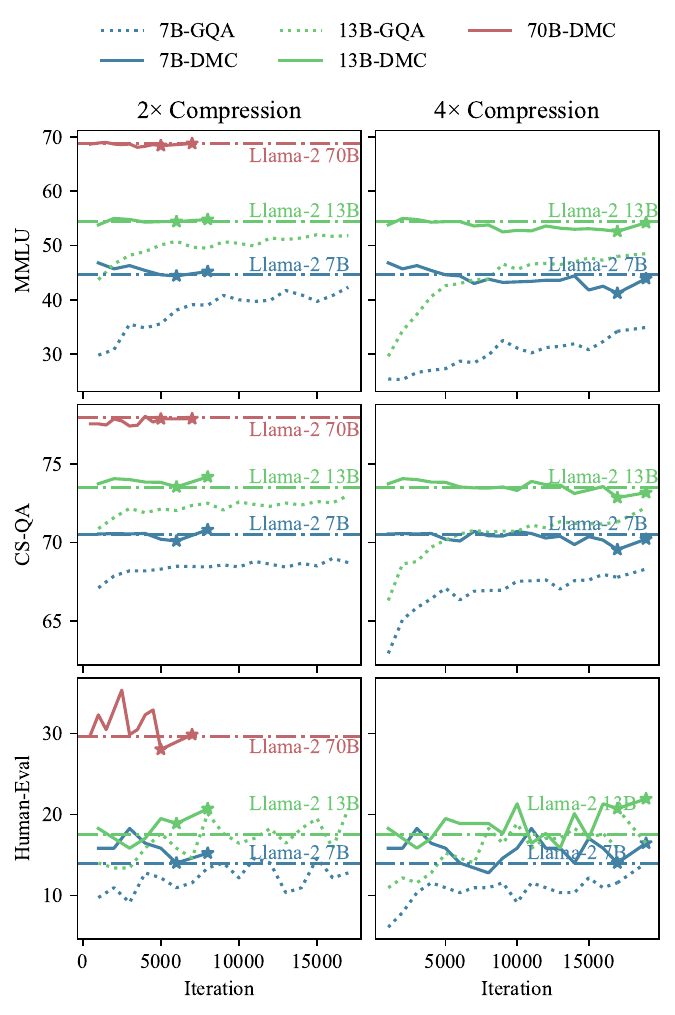}
    \vspace{-1.50em}
    \caption{Sample efficiency of \dmc and GQA. Horizontal lines correspond to the performance of the original Llama 2. Every DMC model was trained first with increasing CR, then with constant CR for the last 2K steps (marked with $\star$).}
    \label{fig:sample-efficiency}
\end{figure}

\textbf{70B: \dmc \textit{and} GQA} Many widely adopted LLMs were pre-trained with GQA which leads to the question of whether \textit{\dmc and GQA can be used together to reap compounded benefits}. To investigate this, we retrofit Llama 2 70B, which has been pre-trained with $8\times$ GQA. We further compress its KV cache with \dmc to $2\times$ CR: this is equivalent to a cache $16\times$ smaller than a vanilla LLM with neither GQA nor \dmc. We observe that the performance remains unchanged, and conclude that \dmc and GQA can be easily and successfully combined.

\textbf{Sample efficiency} To shed light on the sample efficiency of \dmc and GQA, we report their performance on MMLU, CS-QA, and HumanEval across retrofitting steps in \cref{fig:sample-efficiency}. First, for a target CR of $2\times$, it emerges how GQA cannot achieve the performance that \dmc obtains after fine-tuning (at 8K steps) even after more than double the amount of fine-tuning (at 17K steps). This applies to both 7B and 13B scales. \Cref{fig:sample-efficiency} also reveals the importance of the fine-tuning phase in \dmc: a limited amount of extra steps with a fixed CR recovers a significant amount of the original performance (especially for higher target CRs such as $4\times$).

\subsection{Throughput and Latency Measurements}
\label{sec:latency}
\begin{figure}[tp]
    \centering
    \begin{subfigure}[b]{1.00\columnwidth}
      \centering
      \includegraphics[width=1.02\columnwidth]{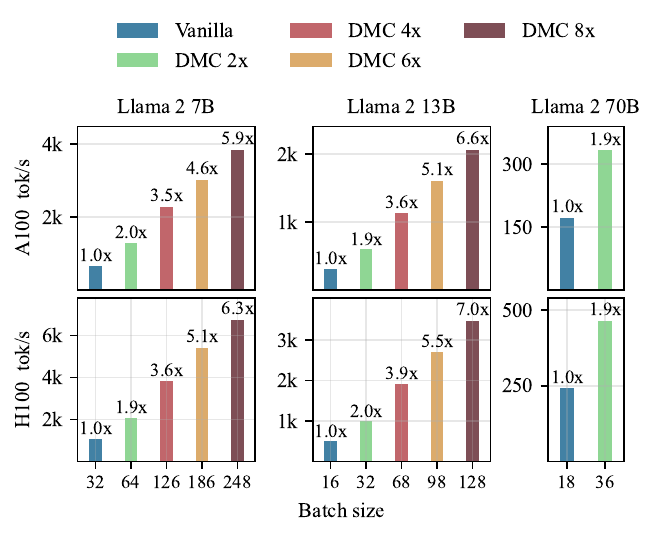}
      \caption{Inference throughput averaged over the generation of the last 1K tokens of a 4K token sequence. On the x-axis, we show the maximum batch size that fits in memory on a single GPU (7B and 13B) or two GPUs with tensor parallelism (70B) for the vanilla LLM, and DMC 2$\times$, 4$\times$, 6$\times$, 8$\times$ models.}
      \label{fig:throughput}
    \end{subfigure}
    \begin{subfigure}[b]{1.02\columnwidth}
      \includegraphics[width=0.96\columnwidth]{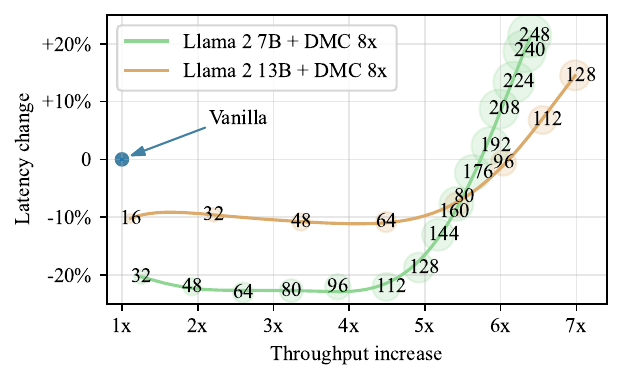}
      \caption{Inference throughput--latency Pareto fronts, charted by varying the batch size with \dmc 8$\times$. Up to a certain batch size, the models are memory-bound and can process additional examples with the same latency. Beyond this point, the models become increasingly compute-bound. Measurements were taken over the generation of the last 1K tokens of a 4K token sequence on an NVIDIA H100.}
      \label{fig:throughput_latency_pareto}
    \end{subfigure}
    \begin{subfigure}[c]{1.02\columnwidth}
      \includegraphics[width=1.02\columnwidth]{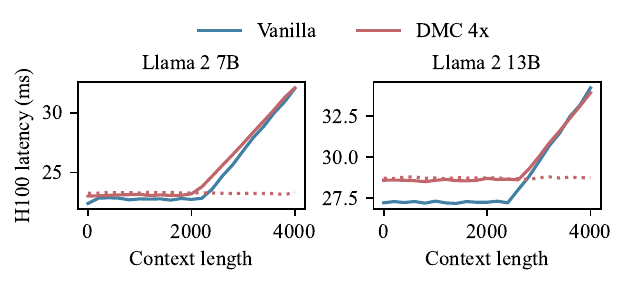}
      \caption{Latency of next token generation. Solid lines denote measurements with the maximum batch size that fits on a single GPU. Dotted lines denote DMC 4$\times$ with the same batch size as the vanilla LLM.}
      \label{fig:latency}
    \end{subfigure}
    \caption{Efficiency measurements with the Megatron-LM framework on NVIDIA A100 80GB and H100 GPUs.}
    \label{fig:efficiency_measurements}
\end{figure}

To verify whether increased CRs result in concrete efficiency gains, we present the performance properties of \dmc in \cref{fig:efficiency_measurements}, estimated within the NVIDIA Megatron-LM framework \citep{Narayanan2021EfficientLL}. Specifically, we run measurements on a single GPU (NVIDIA A100 80GB SXM or H100 SXM) in bfloat16 precision for Llama 2 7B and 13B. For Llama 2 70B, we run the same measurements on two GPUs of the same type with tensor parallelism. We feed the model with 2K tokens of English text and additionally generate 2K tokens in an auto-regressive manner to ensure that the model properly compresses its own generations. We limit the sequence to 4K tokens to avoid issues with context length extrapolation, as this is the maximum length observed by Llama 2 during pre-training. The reported throughput consists of the average over the last 1K tokens.
To maximize the GPU utilization, we increase the batch size to the maximum that fits into memory (see \cref{ssec:memory_compute_bound} for details). Our implementation uses PagedAttention \citep{Kwon2023EfficientMM} with a page size of $32$ tokens.

The compression of the KV cache with \dmc allows for substantial increases in batch size and thus significant throughput gains. As shown in \cref{fig:throughput}, \dmc allows for nearly linear scaling of the batch size with CR, which translates into effective increases in tokens per second compared to the original LLM. Specifically, \dmc enables up to a $2.0\times$ throughput increase for \dmc $2\times$, and up to a $3.9\times$ increase for \dmc $4\times$.
However, the scaling of throughput with CR is not linear. After reaching certain batch sizes, the models start to become compute-bound (\cref{fig:throughput_latency_pareto}), and the latency of inferring the next token begins to increase. Indeed, for \dmc $6\times$ and \dmc $8\times$, throughput starts increasing sub-linearly with CR, reaching up to $5.5\times$ and $7\times$, respectively.

In addition, \cref{fig:throughput_latency_pareto} shows that increasing the batch size with \dmc might trade off latency for throughput, and the Pareto front dominates the vanilla model. 
The different slopes of the curves on the right-hand side of \cref{fig:throughput_latency_pareto} stem mostly from the different hidden sizes of these models. We note that the extra memory saved with \dmc could also be used to cache longer contexts.

Finally, we plot how the latency changes as new tokens are generated with a fixed batch size (\cref{fig:latency}). When we use the largest possible batch size for either model, after generating approximately 2200 tokens, the inference time begins to scale linearly with the context size due to the increasingly dominant cost of reading the KV cache from HBM. On the other hand, if we choose to maintain the same batch size for \dmc $4\times$ as for the original LLM, the memory footprint of the KV cache is reduced and latency for longer contexts significantly improves.

While we acknowledge that the behavior of LLMs at inference time depends on a multitude of factors and implementation details, our measurements in \cref{fig:efficiency_measurements} offer evidence that \dmc can increase throughput and reduce the latency of auto-regressive generation with LLMs. We speculate that in the future, \dmc might be used to grow the KV cache sub-linearly, which would provide an alternative between vanilla Transformers and State Space Models, where memory is constant \citep{fu2023hungry,gu2023mamba}.

\subsection{Per-Head Learned Compression Ratios}
\begin{figure}[t]
    \centering
    \includegraphics[width=0.95\columnwidth]{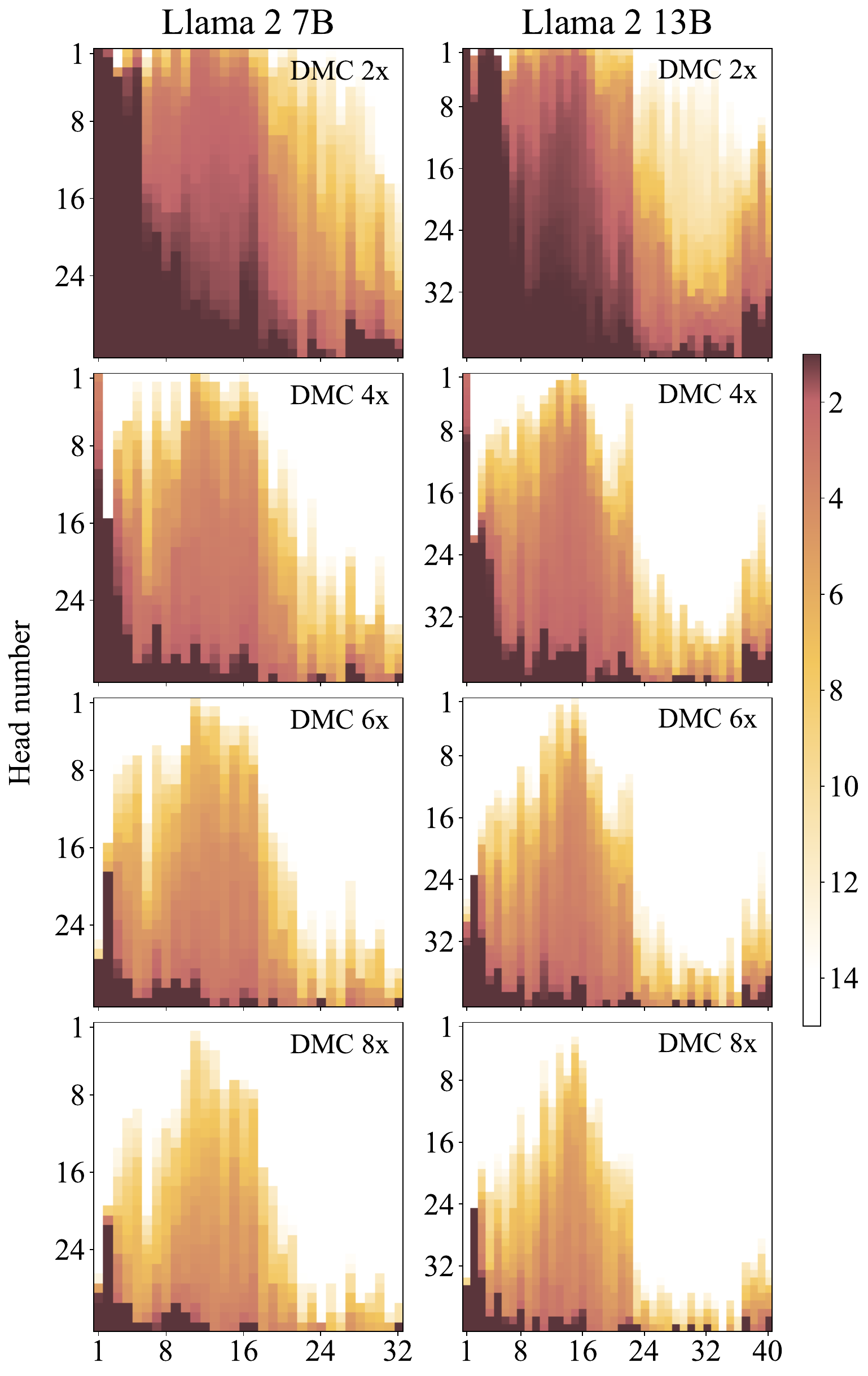}\\
    \noindent\makebox[\columnwidth]{%
        \hspace{9.5mm}
        \includegraphics[width=0.80\columnwidth]{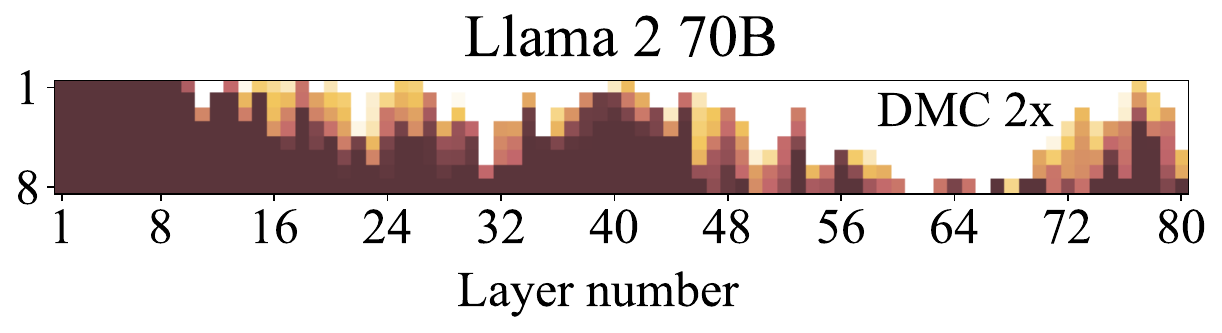}
        \hspace{10mm}
    }
    \caption{Heatmaps of average compression ratios across layers (x-axis) and heads (y-axis). Heads are arranged from the highest compression to the lowest top-down for clarity.}
    \label{fig:cr_heatmap}
\end{figure}

Since the training loss does not enforce any compression schema \textit{a priori}, as it just requires matching a certain \textit{global} CR, we can investigate what schema the model discovers in practice. In  \cref{fig:cr_heatmap}, we report the CR for each layer and head for different scales (7B, 13B, and 70B) and CRs (2$\times$, 4$\times$, 6$\times$, and 8$\times$). From all schema, it emerges how compressing deeper layers ($>$ 16 for 7B, $>$ 22 for 13B, $>$ 44 for 70B) is universally the most popular strategy. However, the very final layers are compressed to a somewhat reduced degree. Fascinatingly, $>4\times$ \dmc achieves extremely high CRs for several heads also in the first few layers. This is arguably counter-productive as token representations are not contextualized yet, which could make decisions (whether to append or accumulate) sub-optimal.

This same pattern of relative preference for compressing certain ranges of layers also emerges during training as we push the model towards the target CR with the auxiliary loss $\ell_{\text{CR}}$. \Cref{fig:ablations-per-layer-compression} in \cref{app:compression_analysis} illustrates how the effective CR increases first in deeper layers, then in some non-contiguous intermediate ranges.

\section{Related Work}

Efficient inference in Transformer models is a subject of extensive research, with detailed overviews provided by several surveys \cite{Pope2022EfficientlyST, Treviso2022EfficientMF}. This section narrows its focus to advancements in Transformer inference efficiency through KV cache size reduction.

Grouped Query Attention \citep[GQA;][]{Ainslie2023GQATG} represents the most widespread strategy, evolving from Multi Query Attention \citep[MQA;][]{Shazeer2019FastTD}. GQA reduces the number of KV heads by allocating shared KV representations across subsets of query heads. Multi-Head Latent Attention \cite{deepseekai2024deepseekv2} compresses the size of the token KV states through a low-rank projection shared across all query heads. Prior efforts in token merging \cite{Zhang2018AcceleratingNT} condensed the entire past context into a single token, while \cite{Liu2018GeneratingWB, Rae2019CompressiveTF} employed strided convolution and mean pooling kernels to reduce the number of KV tokens. Sliding window attention techniques \cite{Beltagy2020Longformer, child2019generating} restrict attention to a maximum of $w$ preceding tokens. Quantization-based approaches \cite{Sheng2023HighthroughputGI, Liu2024KIVIAT, Liu2023LLMQATDQ, Hooper2024KVQuantT1} reduce the KV precision to a smaller number of bits. Though effective in limiting KV cache, such methods perform \textit{fixed-size} compression, unlike the \textit{dynamic} \dmc presented here, which adapts the compression schema based on the input sequence. This adaptation yields superior results, as we prove in an ablation study in \cref{app:dmc-ablations}.

Previous learnable compression methods \citep[\textit{inter alia}]{Anagnostidis2023DynamicCP} decide which tokens to discard from the KV cache. \dmc takes a different approach; instead of dropping tokens, it merges them, thus preserving cached information more faithfully. Moreover, the \dmc compression mechanism has constant complexity relative to the context length, while the method proposed by \citet{Anagnostidis2023DynamicCP} has linear complexity. \citet{Mu2023LearningTC} instead compresses prompts through a costly generation which limits their inference benefits. Moreover, their method is applicable only to compressing the model input while DMC compresses the entire sequence \emph{on the fly}, including both the model input and the generated output.

Non-learnable cache eviction strategies \cite{Zhang2023H2OHO, Sheng2023HighthroughputGI, Liu2023ScissorhandsET, Wang2020SpAttenES, Ge2023ModelTY, Oren2024TransformersAM} utilize attention scores or token properties to filter tokens in the KV cache. These approaches, while bypassing additional training, rely on heuristics and lack the ability to learn the compression mechanisms. In contrast, DMC integrates compression into its learning objective in an end-to-end manner, where compression is synergistic to language generation.

Finally, DMC draws inspiration from Dynamic Token Pooling \citep[DTP;][]{Nawrot2022EfficientTW}, which introduces a learnable boundary predictor to merge the representations of groups of tokens in intermediate layers. DMC improves upon this idea by applying it to the KV cache and introducing a continuous relaxation of the pooling decision during training. Moreover, it enables retrofitting pre-trained LLMs with minimal extra steps rather than training language models from scratch.

\section{Conclusions}
We proposed Dynamic Memory Compression, a method to reduce the length of the KV cache in Transformers, which enhances the efficiency of LLMs at inference time. For every new token, \dmc learns end-to-end whether to append its key--value representations to the cache or merge them with the top element in the cache.
We showed how to retrofit LLMs such as Llama 2 at different scales (7B, 13B, and 70B) into efficient \dmc versions with a negligible amount of extra data and without extra parameters.
\dmc LLMs with up to 4$\times$ compression ratios (CRs) retain (or even improve upon) the performance of the original LLM. Higher CRs (6$\times$ and 8$\times$) instead result in limited performance degradation.
For comparable CRs, \dmc has significantly higher downstream performance and sample efficiency than Grouped Query Attention (GQA), a widespread method for KV cache size reduction, as well as key--value eviction policies such as H$_2$O and TOVA.
In practice, we find that \dmc translates to up to 7$\times$ increase in throughput on an H100 GPU.

\section*{Acknowledgements}
The authors would like to thank Mostofa Patwary for sharing the data blend we used to retrofit the models, Szymon Migacz for his assistance with the computing infrastructure, as well as Przemysław Strzelczyk, Daniel Korzekwa, and Bryan Catanzaro for helpful discussions and support in releasing this paper. This work was supported in part by the UKRI Centre for Doctoral Training in Natural Language Processing, funded by the UKRI (grant EP/S022481/1) and the University of Edinburgh, School of Informatics and School of Philosophy, Psychology \& Language Sciences.

\section*{Impact Statement}
Dynamic Memory Compression in Large Language Models (LLMs) like Llama 2 results in better computational efficiency, reducing both operational costs and environmental impact \citep{Patterson2021CarbonEA}. By enabling higher throughput and lower latency, \dmc democratizes access to advanced AI, making state-of-the-art models suitable for a broader range of hardware. This may not only accelerate innovation across diverse sectors but also promote AI development and applications in an environmentally conscious manner.

\bibliography{custom}
\bibliographystyle{icml2024}

\clearpage
\newpage

\appendix
\section*{Appendix}

\section{Replicating the Original Results}
\label{app:reproduce_baseline}

To make sure that our implementation is correct, for each downstream task, we compare the performance reported in the original Llama 2 paper \citep{touvron2023llama} with those obtained from the Hugging Face Hub checkpoints. Furthermore, we evaluate the impact of using our internal data mixture for up-training, acknowledging that variations in data proportions and preprocessing methodologies can influence model behavior. In particular, we up-train the vanilla pre-trained Llama 2 checkpoint for 200 training steps, amounting to 1B tokens, in accordance with the original Llama 2 training schedule. We compute the average and standard deviation of checkpoints after 50, 100, 150, and 200 steps. In our experiments, we replicate the results reported by the Llama 2 paper almost exactly, as shown in \cref{tab:baseline-results}. Furthermore, we observe that retrofitting the Llama 2 checkpoint on our data mixture has little effect on the model's performance on the downstream benchmarks.

\begin{table*}[t]
\resizebox{1.0\textwidth}{!}{%
\centering
\begin{tabular}{|l|l|l|l|l|l|l|l|l|l|}
\hline
 & \multicolumn{3}{c|}{\textbf{CS-QA}} & \multicolumn{3}{c|}{\textbf{MMLU}} & \multicolumn{3}{c|}{\textbf{Human-Eval}} \\ \cline{2-10} 
 & \textbf{7B} & \textbf{13B} & \textbf{70B} & \textbf{7B} & \textbf{13B} & \textbf{70B} & \textbf{7B} & \textbf{13B} & \textbf{70B} \\ \hline
Paper & 70.6 & 73.7 & 78.5 & 45.3 & 54.8 & 68.9 & 12.8 & 18.3 & 29.9 \\ \hline
Checkpoint & 70.6 & 73.7 & 78.5 & 45.7 & 55.1 & 69.1 & 13.4 & 17.7 & 30.5 \\ \hline
Up-trained & 70.5 ± 0.2 & 73.5 ± 0.1 & 78.0 ± 0.1 & 44.6 ± 0.6 & 54.5 ± 0.3 & 68.8 ± 0.3 & 14.0 ± 1.2 & 17.5 ± 1.5 & 29.6 ± 1.6  \\ \hline
\end{tabular}
}
\caption{Replicating the original up-training results.}
\label{tab:baseline-results}
\end{table*}

\section{Retrofitting Data}
\label{app:retrofitting_data}

The retrofitting corpus comprised sections of The Pile, including BookCorpus2, Books3, Pile-CC, Gutenberg (PG-19), NIH ExPorter, OpenWebText2, Stack Exchange, and Wikipedia (en). Additional datasets included in the corpus were ArXiv, Pushshift Reddit, mC4 (multilingual C4), Common Crawl (CC) dumps from 2017-2023, CC-News, PubMed Central, BigScience Workshop datasets, and The Stack dataset (BigCode project).

\section{Analysis of the Compression Schema Learned by DMC}
\label{app:compression_analysis}

\begin{figure}[ht]
    \centering
    \includegraphics[width=\columnwidth]{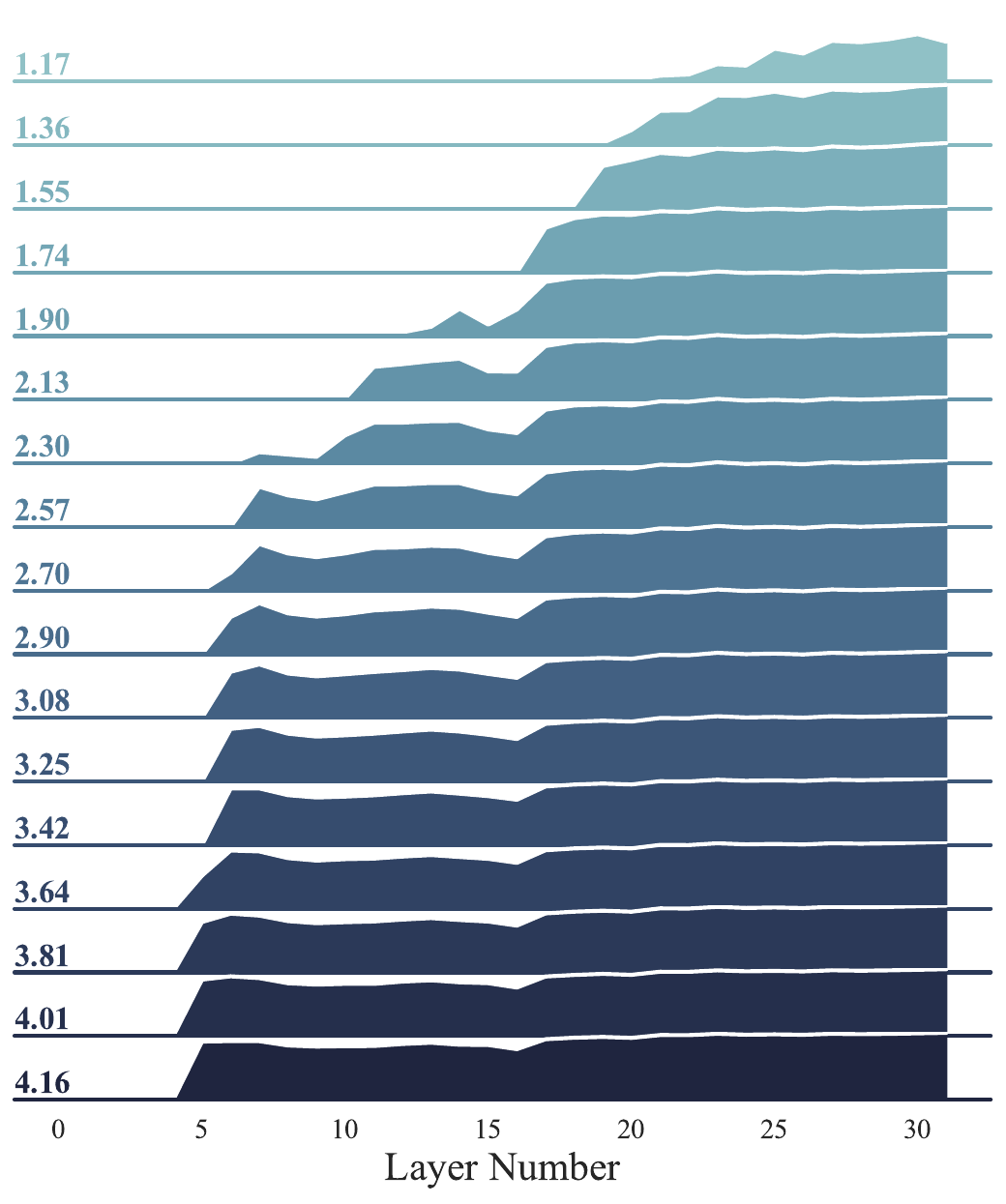}
    \caption{Compression distribution across layers at different stages of retrofitting a Llama 2 7B model. We adhere to the convention where, for a given subplot, a larger space above a given layer indicates greater compression at that layer.}
    \label{fig:ablations-per-layer-compression}
\end{figure}

\paragraph{Evolution throughout the training}
In \Cref{fig:ablations-per-layer-compression}, we illustrate how the CR changes for each layer of the Llama 2 7B model throughout the training from 1$\times$ up to 4$\times$ global CR. Each subplot corresponds to a different global CR which occurs at different stages of the training, going from the smallest (1.17) at the top to the highest (4.16) at the bottom. There is a clear trend such that, for a smaller global Compression Ratio (i.e. at the beginning of the training), the model emphasizes compression in the later layers. As the global Compression Ratio increases, the model keeps on compressing in the final layers but also starts to compress the earlier layers. 
We hypothesize that the token representations in the initial layers do not contain sufficient information to perform any meaningful grouping. Conversely, token representations in the subsequent layers are more defined and, possibly, after several attention layers, already contain redundant/shared information. 

\begin{figure}[h]
  \centering
  \begin{minipage}[b]{0.48\textwidth}
    \centering
    \includegraphics[width=\textwidth]{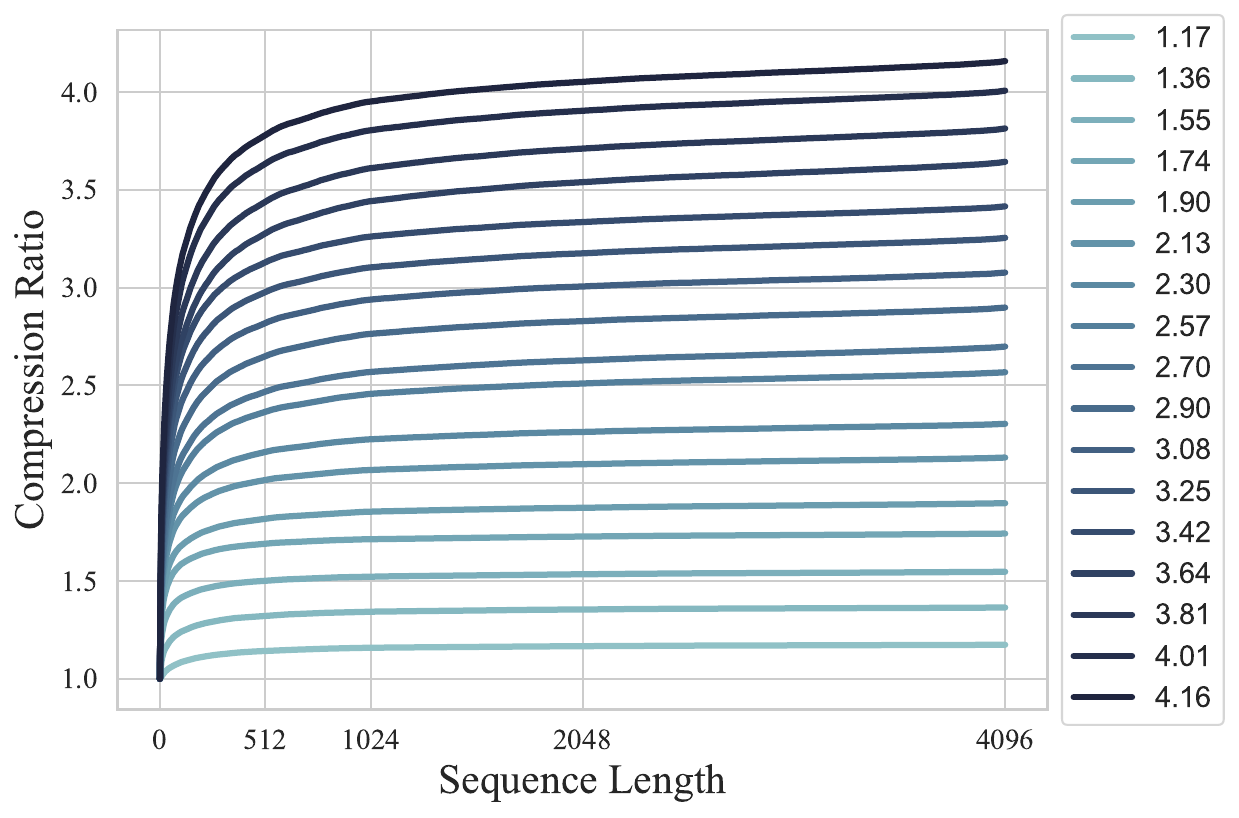}
    \caption{CR achieved by Llama 2 7B for particular sequence lengths across various global CRs.}
    \label{fig:ablations-per-seqlen-compression} 
  \end{minipage}
\end{figure}

\paragraph{Sequence Length versus Compression Ratio}
Do \dmc models compress sequences with a uniform CR independent from their total length? We find that this is not the case. As shown by \cref{fig:ablations-per-seqlen-compression}, the CR increases logarithmically as we increase the total sequence length. This holds true across all global CRs (including both 2$\times$ and 4$\times$).

\begin{figure}[h]
  \centering
  \begin{minipage}[b]{0.48\textwidth}
    \centering
    \includegraphics[width=\textwidth]{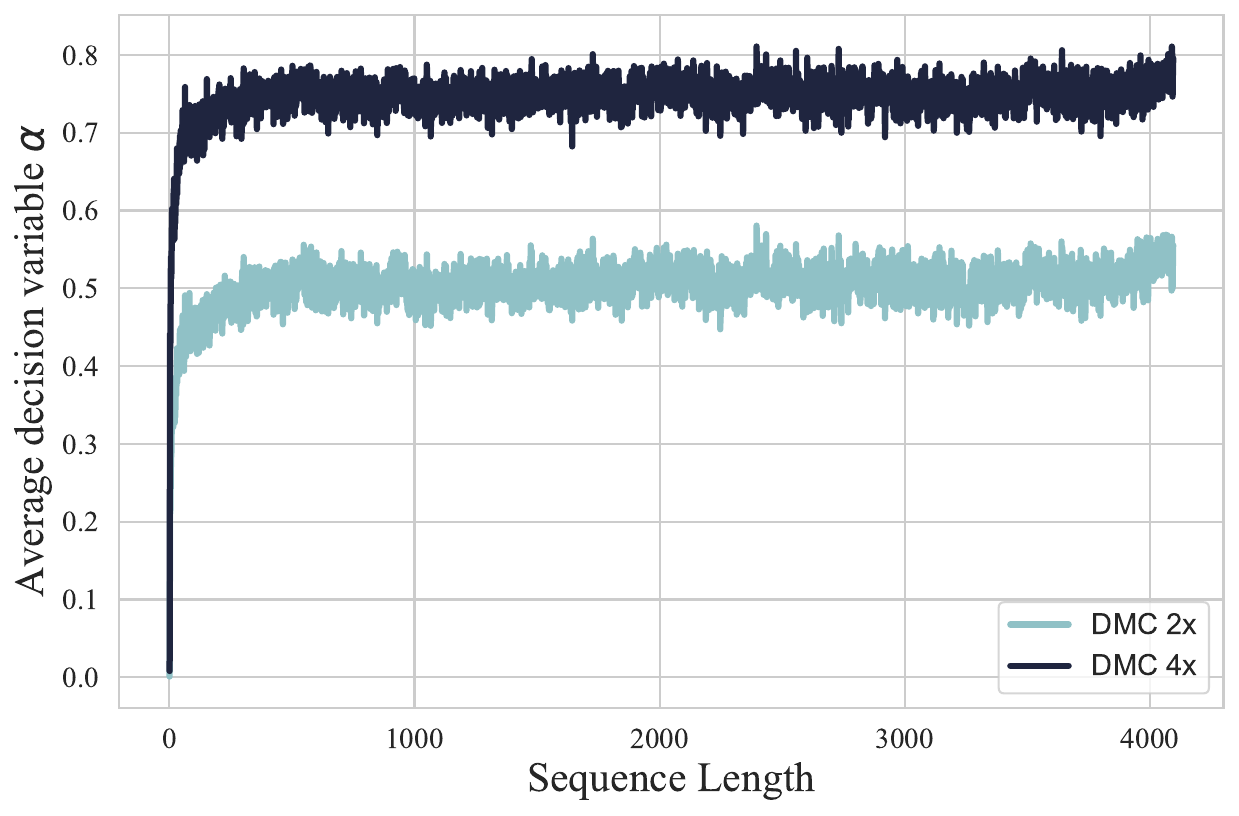}
    \caption{Average value of the decision $\alpha$ for positions (0, 4096) averaged over 128 samples, heads, and layers.}
    \label{fig:ablations-decision-vs-position}
  \end{minipage}
\end{figure}

\paragraph{Absolute Position versus Compression Decision}
Do \dmc models learn a fixed compression schema, or do they exhibit position biases? In \cref{fig:ablations-decision-vs-position}, we plot the average value of the decision variable $\alpha$ across positions in the sequence (0 to 4096). Our observations reveal that the average value of the decision variable $\alpha$ is independent of a token's position in the input sequence which demonstrates that the model does not follow some fixed pattern. This persists across both 2$\times$ and 4$\times$ compression ratios (CR).

\begin{figure}[h]
  \centering
  \begin{minipage}[b]{0.48\textwidth}
    \centering
    \includegraphics[width=\textwidth]{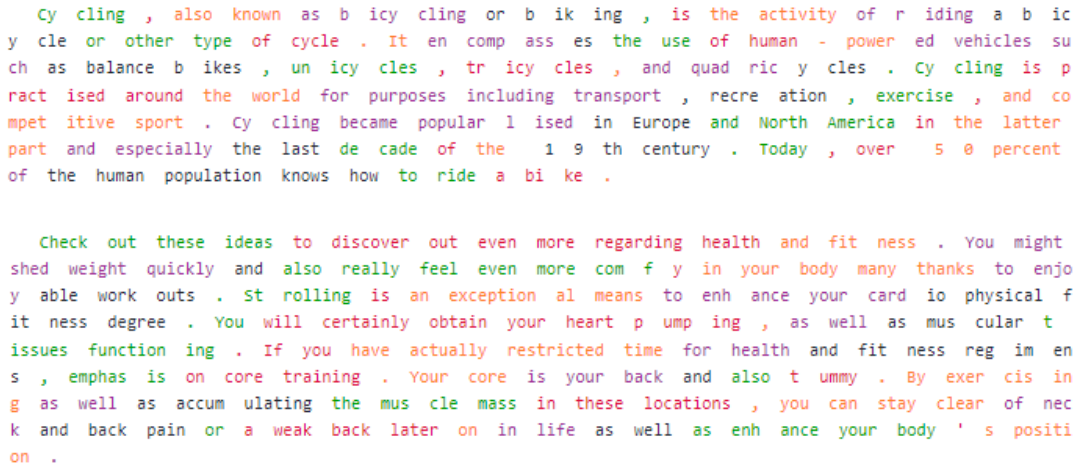}
    \caption{Compression schema found by Llama 2 13B DMC 4$\times$ in layer 0, head 14. Tokens that are merged in the KV cache are marked with the same color.}
    \label{fig:ablations-interpretability_0_14}
  \end{minipage}
\end{figure}

\paragraph{Interpretability}
A natural question arises, whether the compression schema that the model learns is somehow aligned with human intuitions about text segmentation. We analyzed the outputs of Llama 2 13B \dmc with CR $4$ and noticed that some heads compress according to the boundaries of linguistic units, such as words or syntactic phrases. \Cref{fig:ablations-interpretability_0_14} shows the compression schema learned by head 14 in layer 0. In this case, the model merges the subwords back into words \emph{reverting} the tokenizer. Interestingly, some groupings of tokens correspond to semantic units, e.g., ``\textit{1 9 th century}''', ``\textit{5 0 percent}", or ``\textit{a week back later}''. Yet, we also stress that many heads and layers are not interpretable as their behavior does not overlap with linguistic units. 

\begin{figure}[h]
  \centering
  \begin{minipage}[b]{0.48\textwidth}
    \centering
    \includegraphics[width=\textwidth]{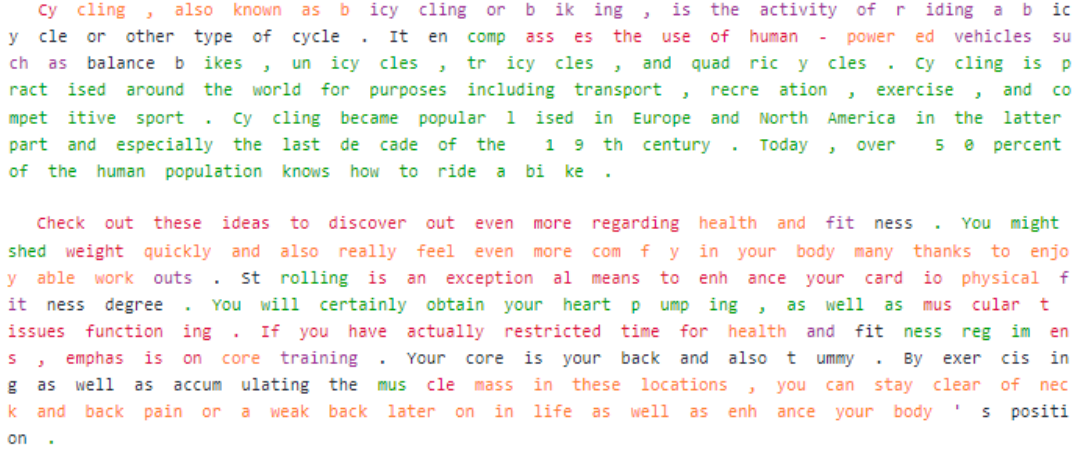}
    \caption{Compression schema found by Llama 2 13B DMC 4$\times$ for layer 24, head 2. Tokens that are merged in the KV cache are marked with the same color.}
    \label{fig:ablations-interpretability_24_2}
  \end{minipage}
\end{figure}

More generally higher layers merge longer token sequences, in line with \cref{fig:cr_heatmap}. For instance, \cref{fig:ablations-interpretability_24_2} shows the decisions of layer 24 head 2.  We leave a more in-depth analysis of compression schemata learned by \dmc to future work.


\section{Similar Per-layer Compression Rates}

In order to alleviate the problem of efficiently storing and retrieving the KV cache of variable-length sequences, we also study a Constrained variant of our algorithm (DMC-C) in which we force all heads in a given layer to maintain a similar compression ratio. A similar compression ratio allows us to store key and value sequences na\"ively as padded tensors with minimal padding. To this end, we add an extra loss term to \cref{eq:cr_loss}
\begin{align} \label{eq:head_loss}
    \ell_{\text{H}} = \sum_{l=1}^{n_l} \sum_{h=1}^{n_h} \sum_{t=1}^{n} \left| \alpha_{lht} -  \sum_{h=1}^{n_h} \alpha_{lht} \right|.
\end{align}

\begin{table}[t]
\centering
\resizebox{1.0\columnwidth}{!}{%
\begin{tabular}{c c c c c c}
\toprule
\textbf{Scale} & \textbf{Method} & \textbf{CR} & \textbf{MMLU} & \textbf{CS-QA} & \multirowcell{2}{\textbf{Human}\\ \textbf{Eval}} \\ 
& \\
\midrule
\multirow{7}{*}{7B} & -- & -- & 44.6 & 70.5 & 14.0 \\ \cmidrule{2-6}
 & GQA    & \multirow{3}{*}{2$\times$} & 39.8 & 68.9 & 12.8 \\
 & DMC & & {45.2} & {\bf 70.8} & \textbf{15.2} \\
& DMC-C    & & \textbf{45.5} & 70.6 & 14.6 \\ 
 \cmidrule{2-6}
 & GQA & \multirow{3}{*}{4$\times$} & 34.7 & 68.3 & 14.0 \\
 & DMC & & {\bf 43.9} & {\bf 70.2} & \textbf{16.5} \\
 & DMC-C & & 38.2 & 69.6 & 14.6 \\ 
\midrule
\multirow{7}{*}{13B} & -- & -- & 54.5 & 73.5 & 17.5 \\ \cmidrule{2-6}
 & GQA    & \multirow{3}{*}{2$\times$}  & 50.2 & 72.7 & 15.9 \\
 & DMC & & {\bf 54.8} & {\bf 74.2} & \textbf{20.7} \\
 & DMC-C    &  & {\bf 54.8} & 73.9 & 18.3 \\ 
 \cmidrule{2-6}
 & GQA & \multirow{3}{*}{4$\times$} & 48.6 & 72.2 & 16.5 \\
 & DMC & & {\bf 54.2} & {\bf 73.2} & \textbf{22.0} \\
 & DMC-C & & 52.4 & 72.9 & 18.3 \\ 
\midrule
\multirow{2}{*}{70B$^*$} & -- & 8$\times^*$ & 68.8 & 78.0 & 29.6 \\ \cmidrule{2-6}
& DMC & 16$\times^*$ & \textbf{68.8} & \textbf{77.9} & \textbf{29.9} \\
& DMC-C & 16$\times^*$ & 67.4 & 78.2 & 31.1 \\
\bottomrule
\end{tabular}
}
\caption{MMLU accuracy, Commonsense Question Answering (CS-QA) accuracy averaged across 6 tasks, and HumanEval Pass@1 for several scales (7B, 13B, and 70B) and compression ratios (CRs; $1\times$, $2\times$, and $4\times$) of Llama 2. Here, we include an extra \dmc variant - \dmc-C, which does not require custom implementation. (*) The 70B model was trained with GQA which compresses the KV cache $8\times$.}
\label{tab:dmc-c}
\end{table}

\cref{tab:dmc-c} compares DMC with its Constrained variant DMC-C. In general, while remaining superior to GQA, DMC-C displays a significant degradation in several configurations, most notably 7B 4$\times$ where it records a drop of -6.4 in MMLU compared to the ceiling. On the other hand, DMC recovers all performance loss in DMC-C. When combined with custom attention implementations that do not require excessive padding, standard DMC should therefore be vastly preferred, as it retains the original LLM performance while fully reaping the advantages in memory efficiency.

\begin{figure}[t]
    \centering
    \includegraphics[width=\columnwidth]{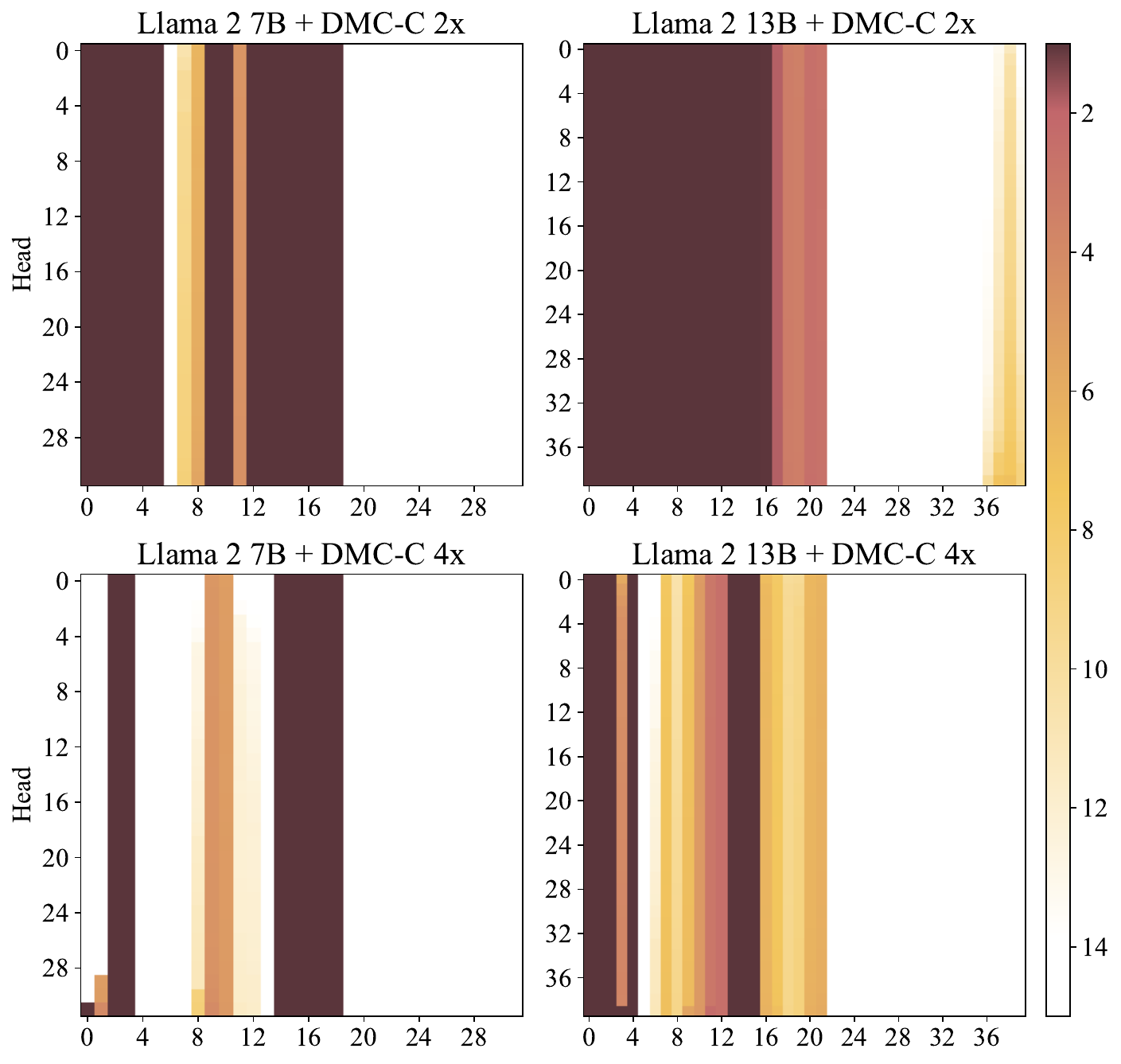}\\
    \noindent\makebox[\columnwidth]{%
        \hspace{9.5mm}
        \includegraphics[width=0.95\columnwidth]{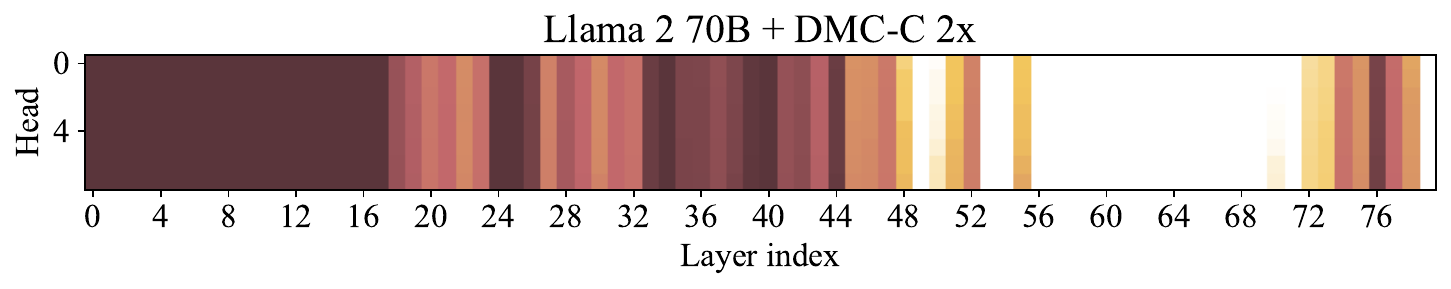}
        \hspace{10mm}
    }
    \caption{Heatmaps of average compression ratios across layers (x-axis) and heads (y-axis) for \dmc-C. Heads are arranged from the highest compression to the lowest top-down for clarity.}
    \label{fig:cr_heatmap_dmc-c}
\end{figure}

\paragraph{Compression Schemata learned by DMC-C} Studying the compression schema in \cref{fig:cr_heatmap_dmc-c} learned by \dmc-C, we find a very different pattern compared to \dmc, due to the auxiliary loss forcing the model to compress similarly across heads in the same layer. Nevertheless, we observe a similar global preference for compressing deeper layers.

\section{Training Ablations}
\label{app:retrofitting_ablations}

\paragraph{Training Steps per Increase in CR}

\begin{figure}[t]
    \centering
    \includegraphics[width=\columnwidth]{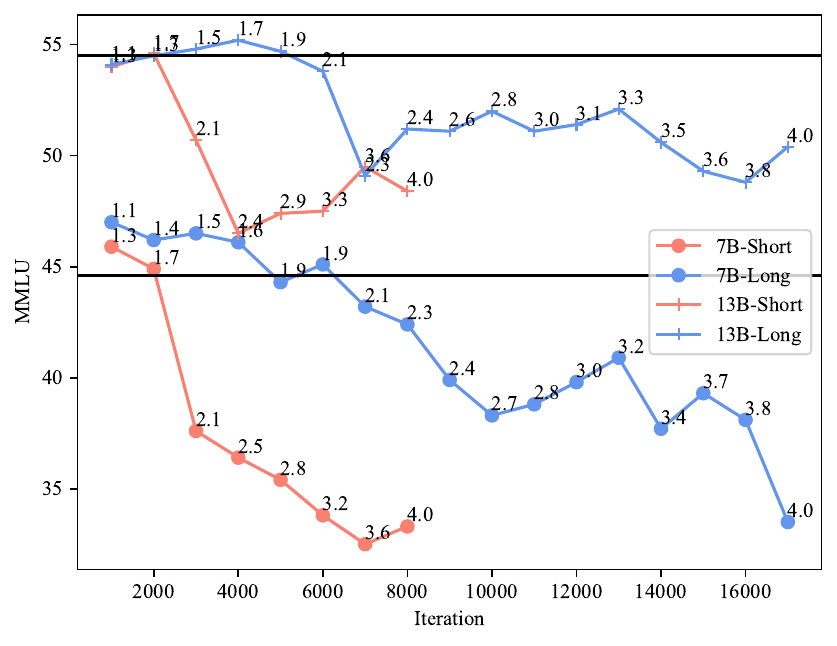}
    \caption{Different up-training regimes for \dmc-C: Short (red) increases CR by 1 every 6K steps, Long (blue) increases CR by 1 every 3K steps. Horizontal lines correspond to the performance of the original Llama 2.}
    \label{fig:regimes}
\end{figure}

Another advantage of \dmc is its high flexibility. In fact, based on the availability of resources, different regimes can be chosen when annealing the CR to the target during up-training. In \cref{fig:regimes}, we compare a \textsc{Short} and a \textsc{Long} regime for the constrained variant of \dmc (\dmc-C), which continuously increases the CR by 1 every 3K and 6K steps (12B and 24B tokens), respectively. Evidently, there exists a trade-off between training steps (hence, time) and performance. Additionally, \cref{fig:regimes} showcases another aspect of the higher flexibility \dmc affords: it is compatible with arbitrary real-valued CRs, as opposed to integer CRs divisible by 2 as in GQA.

\paragraph{Schedules of Target CR}

\begin{figure}[t]
    \centering
    \includegraphics[width=\columnwidth]{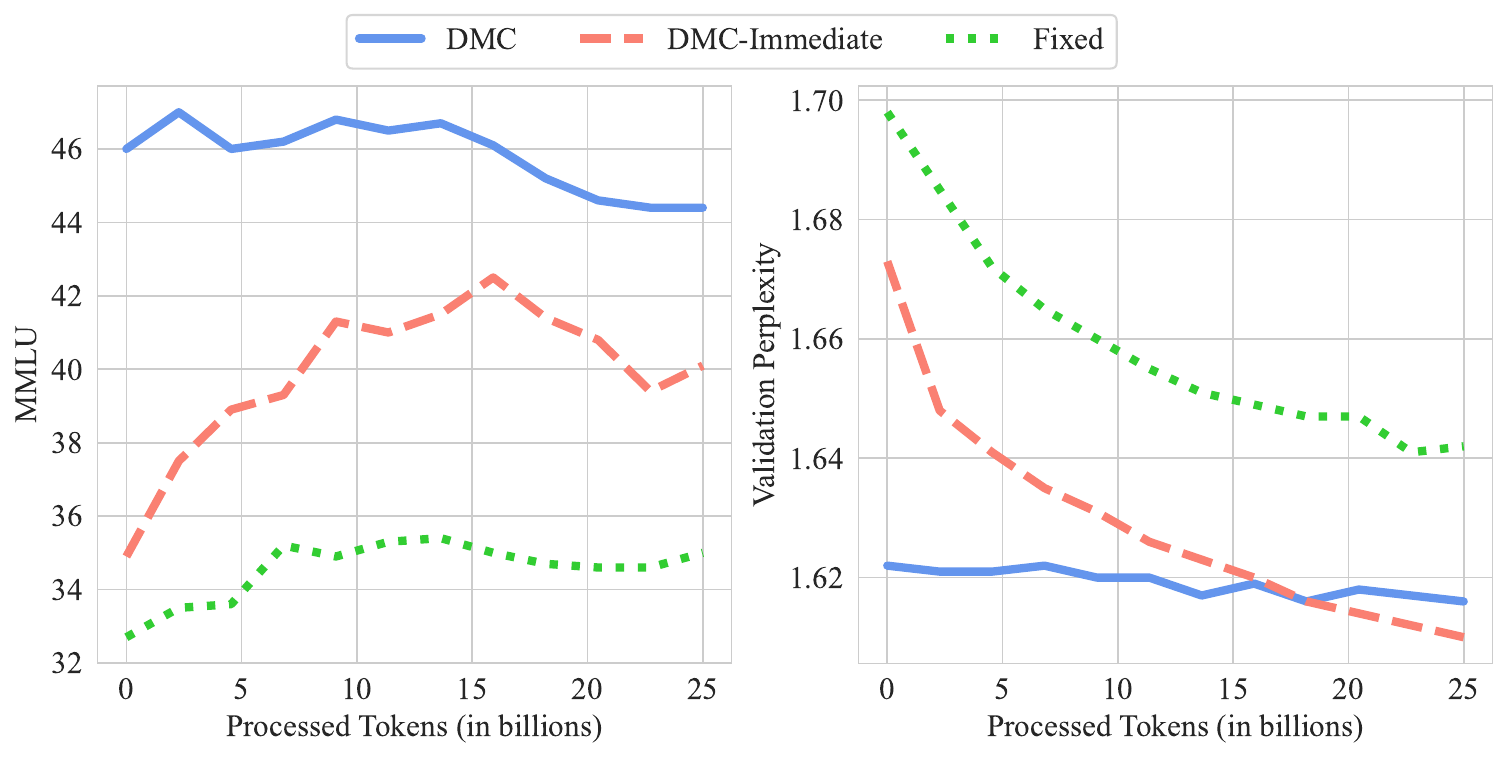}
    \caption{Validation perplexity and MMLU accuracy vs training steps for Fixed Memory Pooling and a variant of DMC where the auxiliary loss CR is immediately set to the target on the onset of training. All models follow the regular training schedule and are trained up to 2$\times$ CR.}
    \label{fig:uptraining_ablations}
\end{figure}

Additionally, we explore how different schedules for the target CR impact the model performance. In the standard setup, this is linearly annealed from 1 to the target CR throughout the duration of training. Here, we compare it with a setup where the CR used in the auxiliary loss for compression is set to the target from the start (DMC-immediate). We show the results in \cref{fig:uptraining_ablations}. As expected, DMC-immediate has a perplexity spike at the beginning when the model quickly increases the CR due to the auxiliary loss. While perplexity is recovered during training, even to a lower point than DMC with annealing, downstream accuracy on the MMLU benchmark is degraded across the board. This showcases why avoiding perplexity spikes is fundamental to successfully retrofitting an LLM.

\section{DMC Ablations}
\label{app:dmc-ablations}

\begin{figure}[t]
    \centering
    \includegraphics[width=\columnwidth]{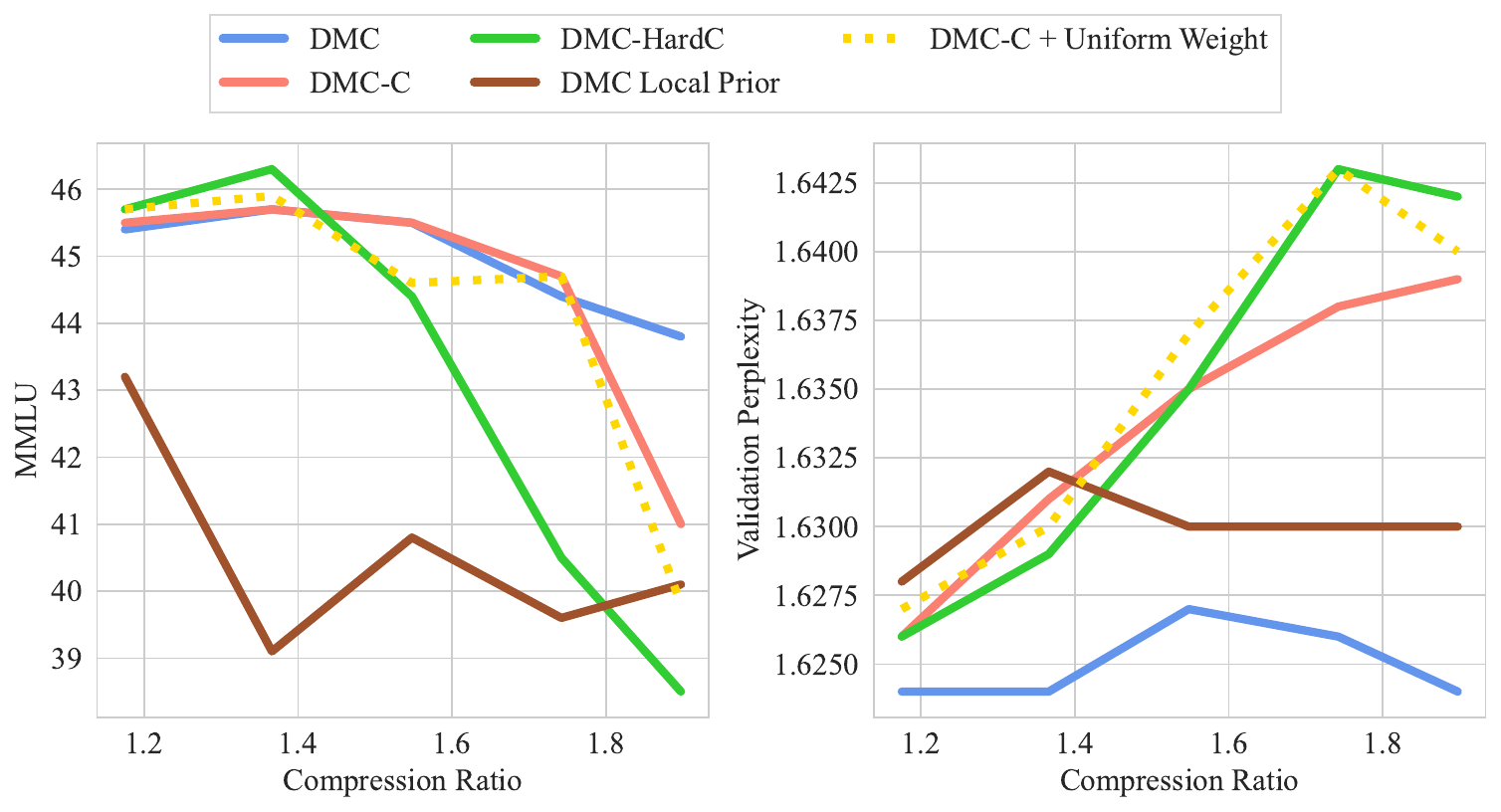}
    \caption{Validation perplexity and test MMLU accuracy vs compression ratio for different types of compression priors, in-layer relaxations, and importance scores. All models follow the \textsc{short} training regime and are trained up to 2$\times$ CR.}
    \label{fig:ablations-dmc-mmlu-pplx}
\end{figure}

\paragraph{Fixed vs Learned Memory Compression}

We assessed the importance of dynamically learning compression decisions in \dmc by comparing it with Fixed Memory Pooling, which reduces the overall number of tokens in memory by deterministically averaging every $n$ tokens, where $n$ in this case is identical to the compression ratio. The results, shown in \cref{fig:uptraining_ablations}, demonstrate that the dynamic component of \dmc is crucial to achieving lower perplexity as well as higher downstream accuracy.

\paragraph{Global vs Local (Layer-wise) Compression Prior}

We compare two approaches to compression: a \textit{Local} Prior, which enforces a pre-specified compression ratio (CR) in each layer independently, requiring every layer to compress approximately the same amount, and a \textit{Global} Prior used by default \dmc, which applies a pre-specified CR across all layers, giving the model the freedom to apply different CRs in each layer, provided that their average compression equals the global CR. \Cref{fig:ablations-dmc-mmlu-pplx} clearly indicates that the Global Prior (DMC in \Cref{fig:ablations-dmc-mmlu-pplx}) improves MMLU performance compared to the Local Prior.

\paragraph{In-layer Relaxation}

We then compare three strategies to determine how similar compression schemata for heads within each layer should be (assuming a global prior):
\begin{enumerate}
    \item {\dmc}: There are no constraints on the decision and importance scores, except for the global loss nudging the model towards a pre-defined CR.
    \item {\dmc-C}: Different heads can have varying decision and importance scores within each layer. However, an auxiliary loss encourages the model to maintain similar CRs among all heads within the layer.
    \item {\dmc-HardC}: Decision scores $\alpha_t$ and importance scores $\omega_t$ are shared across heads, leading to the same shortening schema within each layer across heads.
\end{enumerate}

As per \cref{fig:ablations-dmc-mmlu-pplx}, the default \dmc strategy shows a consistent MMLU performance across varying CRs, while both \dmc-C and \dmc-HardC exhibit a sharp drop in MMLU as the compression reaches 1.9$\times$. Moreover, in \cref{tab:dmc-c} we report a more thorough comparison between DMC and DMC-C. In general, while remaining superior to GQA, \dmc-C displays a significant degradation in several configurations when compared to regular \dmc.

\paragraph{Importance Scores}
Finally, we assess the impact of predicting importance scores for accumulation as opposed to uniformly weighting each token in a group. \cref{fig:ablations-dmc-mmlu-pplx} shows that \dmc-C with Uniform Weighting is worse than learned weighting \dmc-C.

\section{Masking Implementation Details}
\label{app:masking_details}

We mask the unnormalized attention score for the pair $(i, j)$ as follows:
$$\hat{a}_{(i, j)} = \underbrace{\frac{\qq_i[1:d_h]^\top \kk_j[1:d_h]}{\sqrt{d_h}}}_{\text{attention score}} + \underbrace{\log(1 - \alpha_{j+1}) \vphantom{\frac{1}{\sqrt{d_h}}}}_{\text{attention mask}}.$$
We rely on the memory-efficient implementation of MHSA from PyTorch, which allows adding arbitrary masks to the attention scores before softmax. Notwithstanding this, at inference time \dmc remains compatible with efficient libraries for attention such as Flash Attention \citep{dao2022flash}. The $\log(1 - \alpha_{j+1})$ term is calculated as $\text{log-sigmoid}(-\alpha_{j+1})$ for better numerical precision.

\section{Limitations}
This paper is focused on retrofitting existing LLMs into \dmc variants. In our preliminary experiments with \textit{pre-training LLMs with \dmc from scratch}, we obtained negative results when compared to the training curve of GQA. We speculate that this is due to the mutual dependency of modeling and segmenting data: when token representations are not of sufficient quality, boundary decisions are unreliable. Vice versa, incorrect boundary decisions may lead to poor token representations. This creates a vicious cycle that may be broken by techniques that facilitate convergence, such as an Expectation Maximization-style alternation between modeling and segmenting.

\end{document}